\documentclass[10pt,twocolumn,letterpaper]{article}

\usepackage{wacv}              

\usepackage{graphicx}
\usepackage{amsmath}
\usepackage{amssymb}
\usepackage{booktabs}

\usepackage{colortbl} 
\usepackage[table,xcdraw]{xcolor} 
\usepackage{booktabs} 
\usepackage{multirow} 

\usepackage[pagebackref,breaklinks,colorlinks]{hyperref}

\usepackage[capitalize]{cleveref}
\crefname{section}{Sec.}{Secs.}
\Crefname{section}{Section}{Sections}
\Crefname{table}{Table}{Tables}
\crefname{table}{Tab.}{Tabs.}

\usepackage[accsupp]{axessibility} 

\def\wacvPaperID{2242} 
\def\confName{WACV}
\def\confYear{2025}

\begin{document}

\title{Adaptive and Temporally Consistent Gaussian Surfels for Multi-view Dynamic Reconstruction}



\author{
Decai Chen$^{1,2}$ \quad Brianne Oberson$^{1,3}$ \quad Ingo Feldmann$^1$ \\ Oliver Schreer$^1$ \quad Anna Hilsmann$^1$ \quad Peter Eisert$^{1,2}$
\\
$^1$Fraunhofer 
HHI \quad $^2$Humboldt University of Berlin \quad $^3$Technical University of Berlin \\
{\tt\small \{first\}.\{last\}@hhi.fraunhofer.de} 
}


\twocolumn[{%
\renewcommand\twocolumn[1][]{#1}%
\maketitle
\centering
\includegraphics[trim={0.2cm 0.5cm 0.2cm 0.5cm},clip,width=.99\linewidth]{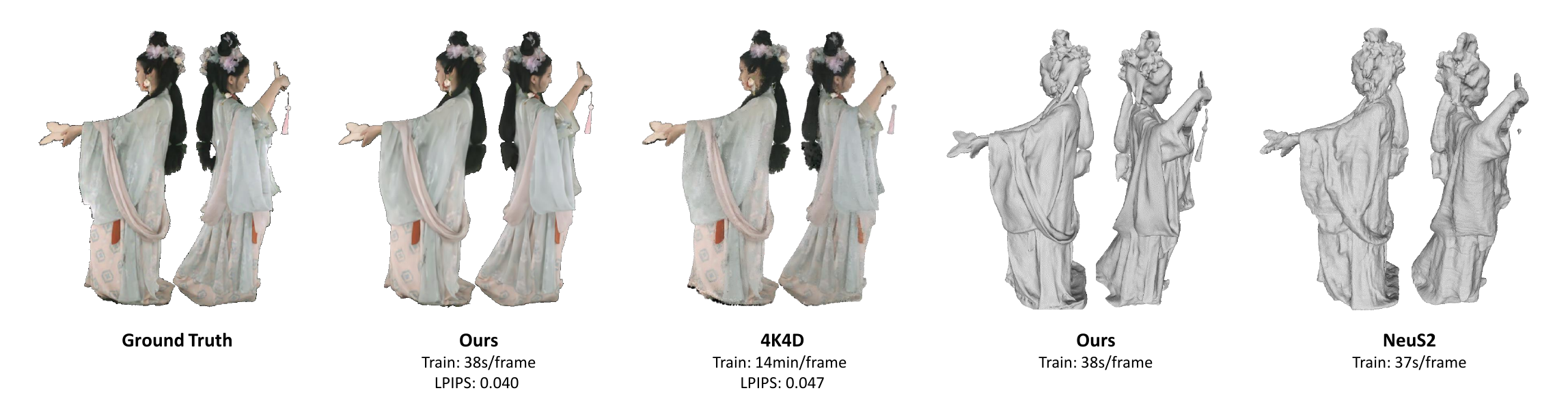}
\captionof{figure}{Comparison of our proposed method on a scene from the DNA-Rendering dataset \cite{2023dnarendering}. The training time and LPIPS scores (lower is better) are averaged across the sequence. Our approach not only achieves photorealistic novel view rendering with significantly reduced training time compared to the recent method \cite{xu20244k4d}, but also produces finer surface meshes, surpassing the state-of-the-art results \cite{wang2023neus2}.}
\label{fig:teaser}
\vspace{2em}
}]

\begin{abstract}

3D Gaussian Splatting has recently achieved notable success in novel view synthesis for dynamic scenes and geometry reconstruction in static scenes. Building on these advancements, early methods have been developed for dynamic surface reconstruction by globally optimizing entire sequences. However, reconstructing dynamic scenes with significant topology changes, emerging or disappearing objects, and rapid movements remains a substantial challenge, particularly for long sequences. To address these issues, we propose AT-GS, a novel method for reconstructing high-quality dynamic surfaces from multi-view videos through per-frame incremental optimization. To avoid local minima across frames, we introduce a unified and adaptive gradient-aware densification strategy that integrates the strengths of conventional cloning and splitting techniques. Additionally, we reduce temporal jittering in dynamic surfaces by ensuring consistency in curvature maps across consecutive frames. Our method achieves superior accuracy and temporal coherence in dynamic surface reconstruction, delivering high-fidelity space-time novel view synthesis, even in complex and challenging scenes. Extensive experiments on diverse multi-view video datasets demonstrate the effectiveness of our approach, showing clear advantages over baseline methods.
Project page: \url{https://fraunhoferhhi.github.io/AT-GS}

\end{abstract}


\section{Introduction}
\label{sec:intro}
Recovering dynamic scenes with high fidelity from multi-view videos presents a significant challenge in computer vision and graphics, with applications spanning virtual reality, cinematic effects, and interactive media. While many existing methods focus on creating visually appealing and immersive representations of dynamic environments, they often fall short when integrated in modern graphics engines, which require precise and temporally stable surface meshes for tasks such as geometry editing, physics-based simulations, animation, and texture mapping. Therefore, our goal is to develop a method that not only delivers photorealistic rendering of dynamic scenes but also ensures the reconstruction of geometrically accurate and temporally consistent surfaces.

In recent years, the rise of Neural Radiance Fields (NeRF) has gained considerable attention for their powerful ability to achieve photorealistic free-viewpoint rendering using compact volumetric representation and differentiable alpha composition \cite{nerf, muller2022instant, chen2022tensorf}.
Building on this foundation, numerous subsequent works \cite{hypernerf, d-nerf, cao2023hexplane, Kplanes, shao2023tensor4d, DyNeRF, Im4D} 
have further explored the synthesis of free-viewpoint videos for dynamic scenes. While NeRF-inspired approaches have driven significant progress, they often struggle with inefficiencies in training time and rendering speed. In contrast, the recent introduction of 3D Gaussian Splatting (3DGS) \cite{3dgs} marks a significant transition towards explicit point-based representations using differentiable rasterization, which offers more efficient training and high-fidelity real-time rendering. Recent advancements \cite{luiten2024dynamic3dg, sun20243dgstream, stg, lin2024gaussian, kratimenos2023dynmf, yang2024deformable, 4dgs, huang2024scgs, das2024neural, katsumata2023efficient, gao2024gaussianflow}
have further demonstrated that Gaussian Splatting achieves superior performance in rendering complex, time-varying environments.

Existing surface reconstruction techniques, including those based on multi-view stereo \cite{yao2018mvsnet, chen2021accurate, ding2022transmvsnet, cao2024mvsformerpp}, neural implicit representations \cite{neus, Fu2022GeoNeus, dneus, li2023neuralangelo}, and more recently, 3D Gaussian Splatting \cite{guedon2024sugar, 2dgs, Yu2024GOF, gsurfels}
have proven effective in static scenes.
However, directly adapting them per-frame to time-varying real-world scenes presents challenges, such as significantly prolonged training times and temporal inconsistencies across frames. An alternative approach is to reconstruct the entire dynamic sequence within a single holistic model \cite{wang2024vidu4d, mags, DGMesh, chen2023dynamic, choe2023spacetime, ndr, johnson2023unbiased}, such as by deforming a canonical space. However, globally representing dynamic scenes with significant topology changes, emerging or disappearing objects, and rapid movements remains a substantial challenge, particularly for long sequences.

To address these challenges, we propose Adaptive and Temporally Consistent Gaussian Surfels (AT-GS), a novel method for efficient and temporally consistent dynamic surface reconstruction from multi-view videos. Our approach utilizes a coarse-to-fine incremental optimization process based on a per-frame Gaussian surfels representation.

Initially, we train the first frame of the sequence using a standard static multi-view reconstruction technique representing the scene as Gaussian surfels \cite{gsurfels}. For each subsequent frame, we learn the SE(3) transformation
to coarsely align Gaussian surfels from the previous frame to the current one. 
We introduce a unified densification strategy combining the strengths of clone and split. Additionally, we design adaptive probability density function (PDF) sampling, guiding the splitting process using the magnitude of view-space positional gradients.

Another challenge in dynamic reconstruction is maintaining temporal consistency. In dynamic reconstruction, slight temporal jittering between frames caused by the randomness of optimization can lead to visible artifacts, especially in regions with minimal textures, where different Gaussian configurations may produce visually identical results. To address this, we first predict the optical flow between neighboring frames of the same view and warp the rendered normal map from the previous frame to the current one. We then enforce consistency in the curvature maps derived from the normals of consecutive frames. This method indirectly ensures temporal coherence in the rendered depth maps and Gaussian orientations, resulting in more stable and accurate final surface geometry.

In summary, our contributions include:
\begin{itemize}
  \item  A method for efficiently reconstructing dynamic surfaces from multi-view videos using Gaussian surfels.
  \item A unified and gradient-aware densification strategy for optimizing dynamic 3D Gaussians with fine details.  
  \item A temporal consistency approach that ensures stable and coherent surface reconstructions across frames by enforcing consistency on curvature maps.
  \item Extensive experiments that demonstrate our method's advantages including fast training, high-fidelity novel view synthesis, and accurate surface geometry.
\end{itemize}


\begin{figure*}
\begin{center}
\includegraphics[trim={0cm 0cm 0.0cm 0cm},clip,width=0.99\linewidth]{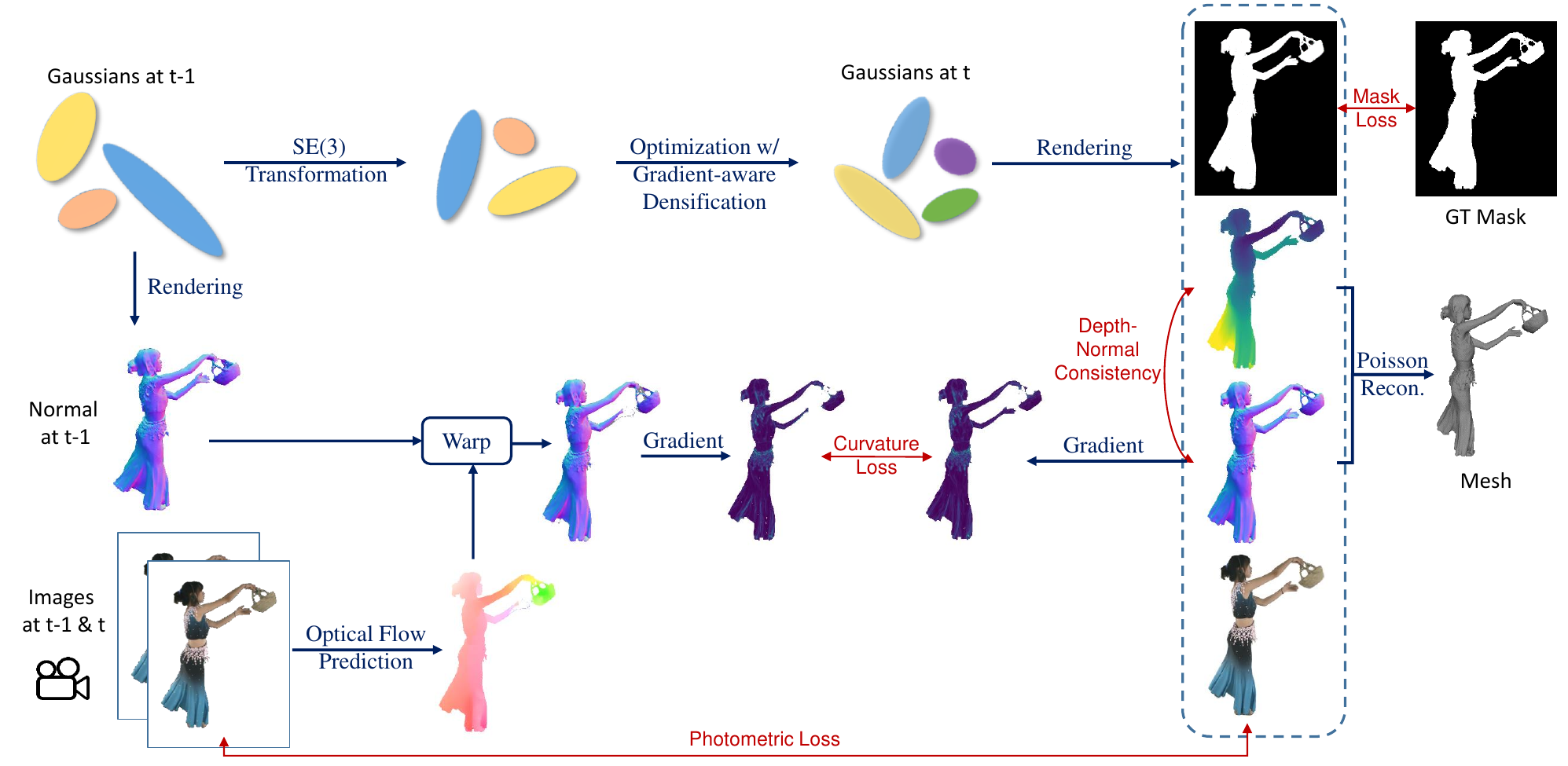}
\end{center}
  \caption{\textbf{Pipeline of Our Method.} Starting with the Gaussian surfels from the previous frame~($t-1$), we first estimate their coarse translation and rotation to align with the current frame~($t$). Subsequently, we optimize all Gaussian attributes, incorporating our gradient-guided densification strategy. For each training view, we render opacity, depth, normal, and color maps (
  from top to bottom in the dashed box) using differentiable tile-based rasterization. Additionally, we predict optical flow between consecutive frames, which warps the rendered normal map from frame~$t-1$ to frame~$t$. We then ensure temporal consistency of the underlying surface by comparing curvature maps derived from the warped and rendered normal maps. Furthermore, we apply photometric loss, depth-normal consistency loss, and mask loss for supervision. Finally, Poisson reconstruction is employed to generate a mesh from the unprojected depth and normal maps.}
\label{pipeline}
\end{figure*}


\section{Related Works}
\subsection{Dynamic Novel View Synthesis}
Recent advancements in dynamic view synthesis have been significantly driven by the development of Neural Radiance Fields (NeRF) \cite{nerf}. Building on this foundation, various methods extend NeRF by conditioning it on temporal states such as frame indices, time coordinates, body poses, or per-frame embeddings, to handle dynamic scenes \cite{nsff, DynamicNeRF, tian2023mononerf, DyNeRF, wang2023mixed, kmh24}. To enhance training and rendering efficiency, recent works factorize the 4D space into lower-dimensional components, such as planes, thereby reducing computational complexity \cite{Kplanes, cao2023hexplane, shao2023tensor4d, icsik2023humanrf, Im4D}. Another line of research focuses on explicitly modeling deformation fields that map the deformed space to a canonical space where the NeRF is embedded \cite{nerfies, hypernerf, NR-NeRF, d-nerf, kirschstein2023nersemble}. Alternatively, some approaches employ incremental training, learning the per-frame differences to achieve efficient dynamic scene representation \cite{song2023nerfplayer, li2022streamrf, ReRF}. Nevertheless, like NeRF, these methods face challenges with slow training and rendering times.

The recently introduced 3D Gaussian Splatting technique \cite{3dgs} dramatically enhances training and rendering speeds through differentiable rasterization. Similar to extending NeRF to dynamic scenes, various approaches predict a deformation field that maps canonical Gaussians to their positions at each observed timestep \cite{yang2024deformable, shaw2023swags, liang2023gaufre, 4dgs, huang2024scgs, jiang2024hifi4g, E-D3DGS, xiao2024bridging, gao2024gaussianflow, wan2024superpoint,guo2024motion}. One alternative direction is to learn temporally continuous motion trajectories of 3D Gaussians via basis functions \cite{katsumata2023efficient, stg, lin2024gaussian, kratimenos2023dynmf}. Besides, other approaches lift 3D Gaussians into 4D by directly incorporating the time dimension \cite{yang2023real, duan20244drotor}. Instead of using Gaussians, 4K4D \cite{xu20244k4d} represents dynamic scenes through point clouds encoded with 4D features, enabling photorealistic and real-time rendering. These methods model entire dynamic scenes using a holistic and temporally smooth representation, which is particularly suitable for monocular video inputs. However, the assumption of accurate cross-frame correspondence often breaks down in complex dynamic scenes with significant topological changes and transient objects. Incrementally optimizing dynamic scenes frame by frame from multi-view videos can mitigate these limitations. Specifically, Luiten \etal \cite{luiten2024dynamic3dg} maintain fixed opacity, color, and size for the optimized 3DGS model from the first frame, and learn per-frame 6-DoF motion of each Gaussian for dense tracking. More recently, 3DGStream \cite{sun20243dgstream} introduces a two-stage training scheme: the first stage trains a Neural Transformation Cache \cite{muller2022instant} for translating and rotating Gaussians with attributes inherited from the first frame, while the second stage spawns and optimizes additional Gaussians using densification and pruning. While relying on the initial Gaussians reduces storage overhead, these methods struggle when subsequent frames deviate significantly from the first frame, such as with emerging objects or topological changes, especially for long sequences. In contrast, our approach allows for the full optimization of per-frame Gaussians including densification and pruning, which enables rapid adaptation to complex dynamic scenes.

Notably, several works guide the movement of Gaussians by supervising the projected Gaussian scene flow with the estimated optical flow of input images \cite{katsumata2023efficient, guo2024motion, gao2024gaussianflow}. However, these approaches maintain a fixed number of Gaussians over time to ensure consistent tracking. In contrast, our method allows flexible densification and pruning, eliminating the need for per-Gaussian correspondence across frames, thereby allowing rapid adaptation to new scenes.

\subsection{Gaussians-based Surface Reconstruction}
Recent advancements in 3D Gaussian Splatting have demonstrated significant progress in surface reconstruction. An earlier work, SuGaR \cite{guedon2024sugar}, introduces regularization terms to better align Gaussians with scene surfaces, leveraging Poisson reconstruction to generate meshes from sampled point clouds. NeuSG \cite{chen2023neusg} and GSDF \cite{yu2024gsdf} integrate 3D Gaussian Splatting with 
SDF 
to jointly optimize these representations. Despite these improvements, challenges such as irregular Gaussian shapes and the presence of artifacts remain. To address these issues, methods like 2DGS \cite{2dgs} and Gaussian-Surfels \cite{gsurfels} flatten 3D volumes into 2D planar disks (i.e., surface elements or surfels) to achieve more precise reconstructions. Additionally, GOF \cite{Yu2024GOF} employs a Gaussian opacity field to facilitate direct geometry extraction. More recently, PGSR \cite{chen2024pgsr} introduced unbiased depth rendering alongside various regularization techniques, while RaDe-GS incorporated a rasterized approach to render depths and surface normals from 3DGS.

Concurrently, several approaches have extended static 3DGS-based surface reconstruction to dynamic scenes from monocular videos. DG-Mesh \cite{DGMesh} proposes cycle-consistent deformation between canonical and deformed Gaussians, mapping these to tracked mesh facets and optimizing Gaussians across all time frames. Vidu4D \cite{wang2024vidu4d} optimizes a bone-based deformation field to transform Gaussian surfels from a canonical state to a warped state, refining rotation and scaling parameters. MaGS \cite{mags} introduces a mutually adsorbed mesh-Gaussian representation, allowing for relative displacement between the mesh and Gaussians, with joint optimization of these elements.
In contrast to these holistic optimization approaches, our method incrementally trains per-frame Gaussian surfels without requiring the entire video sequence during optimization. By employing unified gradient-aware densification and a curvature-based temporal consistency strategy, our method achieves robust dynamic reconstruction, even in complex scenes with significant topological changes.


\section{Method}
In this section, we first introduce our incremental training pipeline for Gaussian surfels tailored to dynamic scenes, as detailed in \cref{subs_incre}. Next, we elaborate on our unified, gradient-guided densification strategy in \cref{subs_dens}, which refines dynamic 3D Gaussians with fine-grained detail. In \cref{subs_temp}, we present our curvature-based temporal consistency approach, ensuring stable and coherent surface reconstructions over time. Finally, we provide details on the model training process in \cref{subs_opti}.

\subsection{Incremental Gaussian Surfels} \label{subs_incre}
 The overview of AT-GS is demonstrated in \cref{pipeline}. Our goal is to accurately reconstruct both the appearance and the geometry of dynamic scenes from multi-view video sequences. The input consists of multi-view RGB image sequences, denoted as ${\mathit{I_{i,j}} : i \in [1,N], j \in [1,M]}$, where $N$ represents the number of frames and $M$ the number of views. Additionally, camera calibration parameters, including intrinsics and poses, are provided.
 
First, we perform full training of Gaussian surfels \cite{gsurfels} for frame~$0$ from a sparse point cloud generated by Structure-from-Motion (SfM) \cite{colmap} or random initialization, following standard static reconstruction practices. For each subsequent frame $t$, we begin with the Gaussians from the previous frame $t-1$ and efficiently adapt them to the current frame using a coarse-to-fine strategy. 

In the coarse stage, only the centers and rotations of the Gaussians are updated. Inspired by 3DGStream \cite{sun20243dgstream}, we train a per-frame Neural Transformation Cache (NTC) \cite{caching, muller2022instant}, consisting of multi-resolution hash encoding and a shallow MLP, which maps spatial positions to SE(3) transformations. Leveraging the spatial smoothness of the voxel-grid representation and linear interpolation, NTC facilitates faster convergence compared to directly optimizing the center and rotation of each Gaussian.

In the fine stage, we refine all learnable parameters (center, rotation, scale, view-dependent color, and opacity) of the Gaussians while allowing for pruning and adaptive gradient-guided densification (\cref{subs_dens}) to capture fine details and accommodate new objects. 
Furthermore, we leverage curvature-based temporal consistency (\cref{subs_temp}) to ensure stable and coherent surface reconstructions over time.

After optimization, we render color, depth, normal, and opacity maps using alpha blending. Similar to \cite{gsurfels}, the rendered depth maps are back-projected and merged in the global 3D space to form a point cloud, with normals 
derived from the rendered normal maps. Lastly, screened Poisson reconstruction \cite{kazhdan2013screened} is applied to generate a surface mesh.

\subsection{Unified and Gradient-aware Densification} \label{subs_dens}

In conventional 3DGS methods \cite{3dgs,gsurfels}, Gaussians with large positional gradients are densified through either cloning or splitting, depending on their sizes. Cloning produces a new Gaussian that retains the size and position of the original, while the original Gaussian shifts in the direction of the positional gradient. On the other hand, splitting generates two smaller Gaussians with positions sampled from the original Gaussian, and then removes the original.
Although effective in static reconstruction, where training occurs from scratch, this conventional densification approach can lead to sub-optimal performance in incremental training of dynamic scenes.

After initializing from the previous frame and applying the SE(3) transformation, our Gaussian representation is already near optimal convergence, and our objective is to optimize it within a limited number of iterations. This presents two challenges. First, due to small gradients and momentum during optimization, the centers of the Gaussians are prone to becoming trapped in suboptimal local minima. Therefore, it is preferable to move the Gaussians rather than preserve them in their original positions, as it is done in conventional cloning. Second, excessive updates to the Gaussians can degrade the results, particularly during splitting, which can displace Gaussians far from their original positions.

To address these challenges, we propose a new adaptive densification strategy for the fine stage of per-frame optimization. This approach unifies the strengths of conventional cloning and splitting into a single step to densify Gaussians with large positional gradients, regardless of their size. Specifically, the original Gaussian moves in the direction of positional gradient, while a new Gaussian of the same size is added by gradient-guided sampling. 

\begin{figure}[t]
\begin{center}
\includegraphics[trim={0cm 0.cm 0.0cm 0.cm},clip,width=0.99\linewidth]{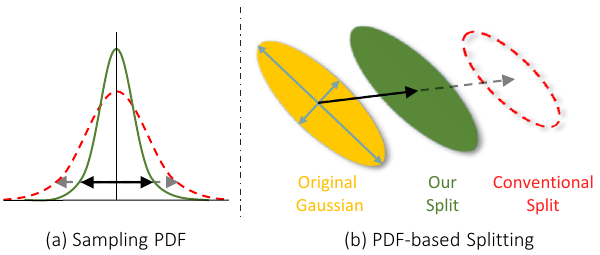}
\end{center}
  \caption{
  \textbf{Gradient-aware splitting.} (a) 1D illustration of sampling PDFs (normal distributions), which determine the positions of new split Gaussians. (b) Conventional splitting (red dashed ellipse) samples a new, smaller Gaussian from a multivariate normal distribution centered at the original Gaussian, with standard deviations equal to its scales. In contrast, our approach (green solid ellipse) adaptively guides the sampling using view-space positional gradients, while preserving the size of the original Gaussian.
      }
\label{fig_dens}
\end{figure}

As illustrated in \cref{fig_dens}, splitting creates a new Gaussian centered at a point sampled from a multivariate normal distribution. Conventional split uses the scales of the original Gaussians as the standard deviations $\boldsymbol{\sigma}$ of the sampling PDFs, which can cause excessive displacement, especially for Gaussians with moderate gradients that require only minor adjustments. This issue is more pronounced in incremental dynamic training scenarios, where the Gaussian representation is already close to convergence and only a few iterations are available to correct positional inaccuracies. 

To overcome this, we utilize the magnitude of the loss function gradients $\|\nabla \mathcal{L}\|$ with respect to view-space positions of the Gaussians to adaptively guide the sampling PDFs. We first clamp $\|\nabla \mathcal{L}\|$ to a maximum of twice their average of all Gaussians being split $\overline{\|\nabla \mathcal{L}\|}$, in order to eliminate outliers. We then normalize these clamped gradients $\|\nabla \mathcal{L}\|$ and use them to scale the standard deviations of their corresponding sampling PDFs:

\begin{equation}
\hat{\boldsymbol{\sigma}} = \boldsymbol{\sigma} \times \frac{\min\left(\|\nabla \mathcal{L}\|, 2 \overline{\|\nabla \mathcal{L}\|}\right)}{2 \overline{\|\nabla \mathcal{L}\|}} .
\label{eqn_split}
\end{equation}


Finally, we initialize the centers of the split Gaussian by sampling from a multivariate normal distribution with standard deviations $\hat{\boldsymbol{\sigma}}$. In this manner, Gaussians with larger gradients are more likely to move further away, allowing significant updates, while those with smaller gradients undergo finer corrections. 
Additionally, instead of shrinking the split Gaussians, we maintain their original size to achieve faster convergence.
Our unified, adaptive densification strategy effectively balances movement and stability, enabling efficient and precise optimization during incremental training for dynamic scenes.

\subsection{Curvature-based Temporal Consistency} \label{subs_temp}

Another key challenge in reconstructing dynamic surfaces is ensuring temporal consistency.
We observe that concatenating per-frame meshes leads to noticeable jittering artifacts during temporal visualization, as shown in the supplementary video.
This temporal inconsistency arises for two main reasons. First, the training process for 3D Gaussian Splatting is susceptible to randomness stemming from GPU scheduling, even with identical input data \cite{gaussian_splatting_issue_89}. Similarly, the inherent randomness in optimization can yield varying results in dynamic reconstruction, even in regions with minimal changes in the input images. Second, low-textured areas lack sufficient multi-view photometric constraints, allowing different Gaussian representations to fit the same input images.
This jittering degrades visual quality, undermining downstream applications like virtual reality, where temporal coherence is crucial.

To overcome this issue, we leverage the local rigidity prior to enhance the temporal consistency of dynamic surfaces. As illustrated in \cref{pipeline}, for each training view of frame $t$, we estimate both forward (from $t-1$ to $t$) and backward (from $t$ to $t-1$) optical flow between consecutive timesteps using an off-the-shelf technique \cite{teed2020raft}. We then compute a confidence mask for the backward flow using cycle consistency. With this masked backward flow, the rendered normal map $\mathbf{N}_{t-1}$ at frame $t-1$ is warped to $\hat{\mathbf{N}}_{t}$ for frame $t$. 

Next, we approximate the curvature map ${\mathbf{C}}$ derived from the normal map ${\mathbf{N}}$ by:
\begin{equation}
\mathbf{C} = \| \left( \left|\nabla_x \mathbf{N}\right| + \left|\nabla_y \mathbf{N}\right| \right) \|_2 .
\label{eqn_curv}
\end{equation}
where $\nabla_x \mathbf{N}$ and $\nabla_y \mathbf{N}$ represent the partial derivatives of $\mathbf{N}$ with respect to the pixel space coordinates $x$ and $y$, while $\|\cdot\|_2$ denotes the Euclidean norm calculated along the spatial (i.e., $xyz$) dimensions. Using \cref{eqn_curv}, we compute the curvature maps for both $\hat{\mathbf{N}}_{t}$ and $\mathbf{N}_{t}$, denoted as $\hat{\mathbf{C}}_{t}$ and $\mathbf{C}_t$, respectively. 

Finally, we define the loss function for temporal consistency as the mean squared error (MSE) between both curvature maps:
\begin{equation}
\mathcal{L}_t = \text{MSE}\left( \hat{\mathbf{C}}_{t}, \mathbf{C}_t \right) .
\label{eqn_l_temp}
\end{equation}

This loss function allows us to maintain local rigidity by ensuring that the orientation of Gaussian surfels is temporally coherent. By supervising the curvature map of the current frame with the warped curvature map from the previous frame, we achieve smoother temporal transitions and more accurate reconstruction of the dynamic scenes.

\begin{figure*}
\setlength{\abovecaptionskip}{2pt} 
\begin{center}
\includegraphics[trim={0cm 0.2cm 0cm 0.2cm},clip,width=0.98\linewidth]{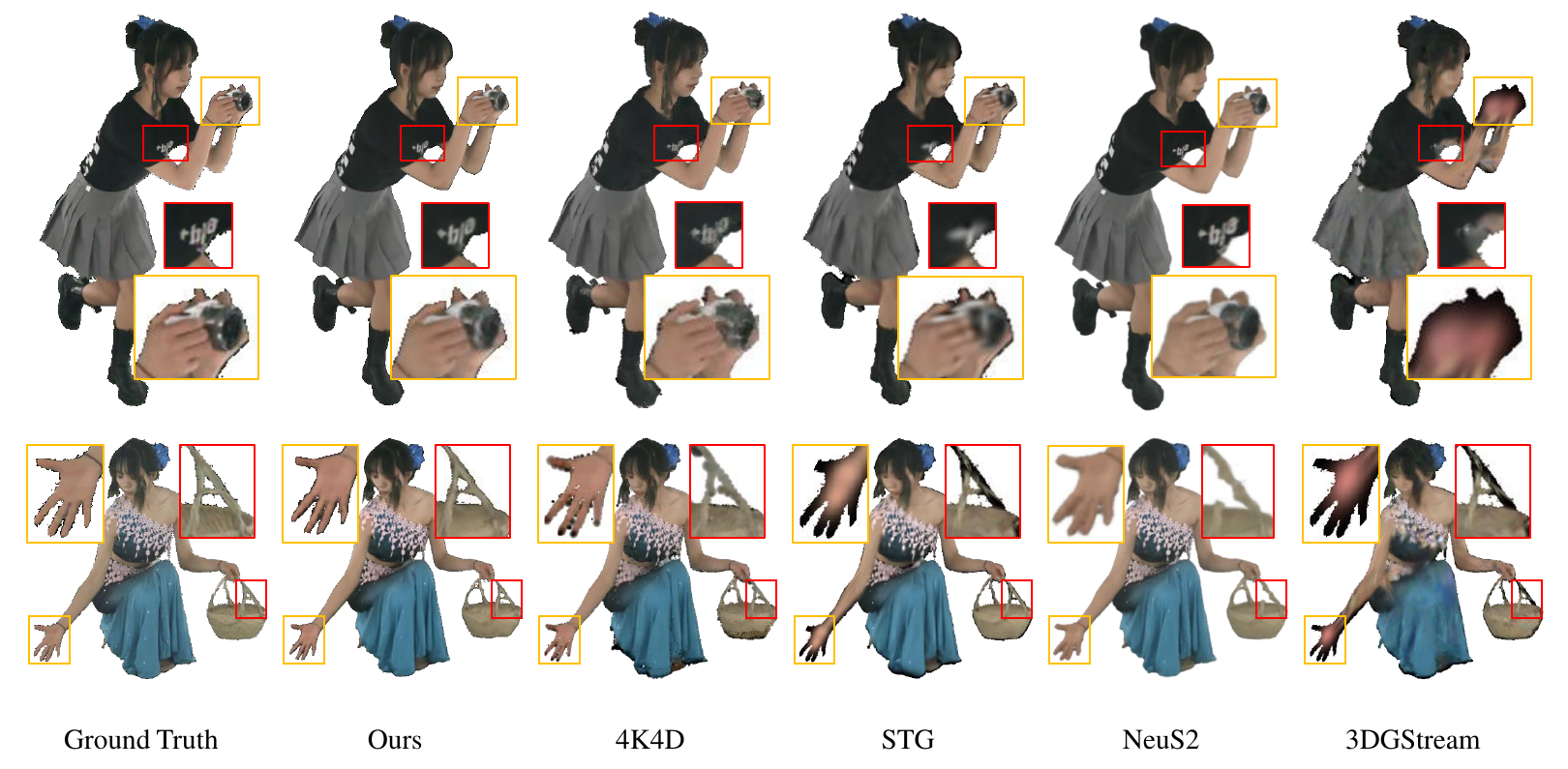}
\end{center}
  \caption{Qualitative comparison of novel view synthesis on the DNA-Rendering dataset \cite{2023dnarendering}.}
\label{fig_dna_compare}
\end{figure*}

\subsection{Optimization} \label{subs_opti}
Our reconstruction process is optimized in an end-to-end manner. Following static reconstruction with Gaussian surfels \cite{gsurfels}, we incorporate a comprehensive set of loss functions to guide the optimization process. In addition to our temporal consistency loss $\mathcal{L}_t$, the per-frame optimization is supervised by several other loss functions: photometric loss $\mathcal{L}_p$, depth-normal consistency loss $\mathcal{L}_c$, opacity loss $\mathcal{L}_o$, and mask loss $\mathcal{L}_m$. The total loss function is thus formulated as:
\begin{equation}
    \mathcal{L} = \mathcal{L}_{p} + \lambda_{t} \mathcal{L}_{t} + \lambda_{c} \mathcal{L}_{c} + \lambda_{o} \mathcal{L}_{o} + \lambda_{m} \mathcal{L}_{m} ,
\label{eqn_loss}
\end{equation}
where $\lambda_{t}, \lambda_{c}, \lambda_{o}, \lambda_{m}$ are weighting factors that balance the contribution of each respective term. 

In contrast to \cite{gsurfels}, our method does not require additional monocular normal prior or its associated loss term, as our framework is capable of supervising the normal maps more effectively through the temporal context. During the coarse stage of per-frame training, we simplify the optimization by focusing solely on the photometric loss $\mathcal{L}{p}$ and mask loss $\mathcal{L}{m}$, which are sufficient for learning the coarse transformation of the Gaussian representation. As the training progresses to the fine stage, we apply the full set of loss terms described in \cref{eqn_loss} to optimize the fine-grained details of the dynamic scene reconstruction.
For all datasets, we train the coarse stage for 200 iterations and the fine stage for 800 iterations. Thanks to our adaptive incremental optimization strategy, we achieve efficient on-the-fly training at approximately 30 seconds per frame. 

\begin{figure*}
\setlength{\abovecaptionskip}{2pt} 
\begin{center}
\includegraphics[trim={0cm 0.5cm 0cm 0.5cm},clip,width=0.98\linewidth]{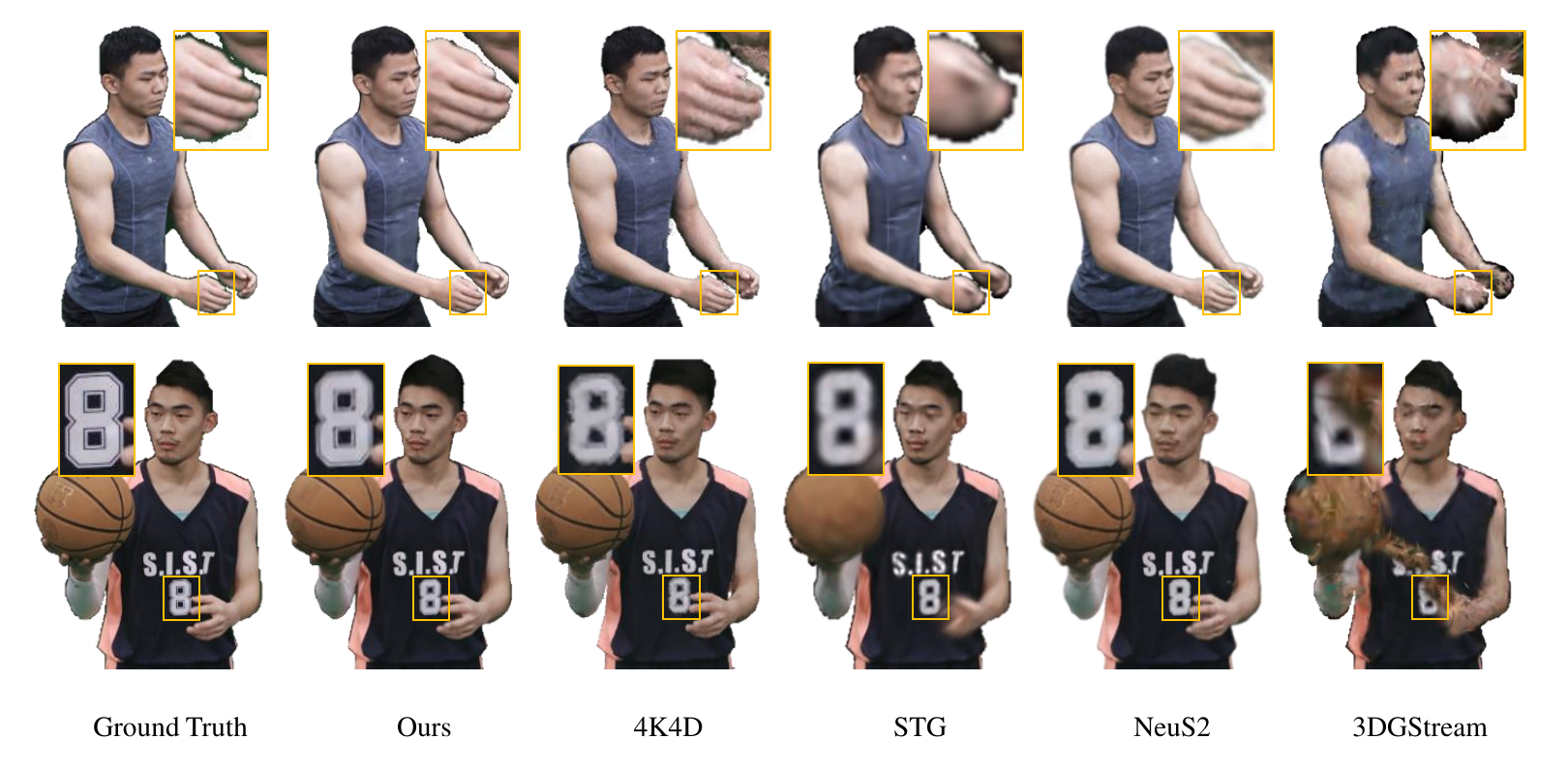}
\end{center}
  \caption{Qualitative comparison of novel view synthesis on the NHR dataset \cite{nhr}.}
\label{fig_nhr_compare}
\end{figure*}

\section{Experiments}

\subsection{Datasets} \label{dataset}
We evaluate our method on the DNA-Rendering \cite{2023dnarendering} and NHR \cite{nhr} datasets for dynamic scene reconstruction. The DNA-Rendering dataset features dynamic scenes of humans with complex clothing, rapid movement, and challenging objects, such as reflective surfaces. It is captured using 48 cameras at a resolution of 2448$\times$2048 and 12 cameras at 
4096$\times$3000. In contrast, the NHR dataset includes 56 cameras capturing three sport scenes and 72 cameras for the basketball scene, with resolutions of 1024$\times$768 and 1224$\times$1024. Our experiments are conducted on five commonly used sequences from DNA-Rendering dataset and all four sequences from the NHR dataset, each containing 150 frames. Following 4K4D \cite{xu20244k4d}, four views are reserved for testing, while the remaining views are used for training. 

\subsection{Comparison} \label{comp}

\begin{table}
\centering
\scalebox{0.9}{ 
\begin{tabular}{llccc}
\toprule 
Type & Method & PSNR$\uparrow$ & SSIM$\uparrow$ & LPIPS$\downarrow$  \\ 
\midrule
\multirow{2}{*}{Holistic} & 4K4D & \cellcolor{yellow!35}34.52 & 0.985 & \cellcolor{yellow!35}0.025 \\ 
 & STG & 28.49 & 0.966 & 0.041 \\ 
\midrule
\multirow{3}{*}{Incremental} & NeuS2 & 33.80 & \cellcolor{yellow!35}0.987 & 0.032 \\ 
 & 3DGStream & 30.78 & 0.974 & 0.047 \\ 
 & Ours & \cellcolor{red!35}35.44 & \cellcolor{red!35}0.988 & \cellcolor{red!35}0.024 \\ 
\bottomrule 
\end{tabular}
}
\caption{Quantitative results on the DNA-Rendering dataset \cite{2023dnarendering}. The best values are highlighted in \colorbox{red!35}{\textbf{red}}, and the second-best values in \colorbox{yellow!35}{\textbf{yellow}}. Metrics are averaged across all scenes.}
\label{tbl_dna}
\end{table}

\begin{table}
\centering
\scalebox{0.9}{ 
\begin{tabular}{llccc}
\toprule 
Type & Method & PSNR$\uparrow$ & SSIM$\uparrow$ & LPIPS$\downarrow$  \\ 
\midrule
\multirow{2}{*}{Holistic} & 4K4D & \cellcolor{red!35}33.65 & \cellcolor{yellow!35}0.972 & \cellcolor{red!35}0.039 \\ 
 & STG & 28.05 & 0.949 & 0.074 \\ 
\midrule
\multirow{3}{*}{Incremental} & NeuS2 & 33.04 & \cellcolor{yellow!35}0.972 & \cellcolor{yellow!35}0.047 \\ 
 & 3DGStream & 30.70 & 0.955 & 0.083 \\ 
 & Ours & \cellcolor{yellow!35}33.55 & \cellcolor{red!35}0.973 & 0.054 \\ 
\bottomrule 
\end{tabular}
}
\caption{Quantitative results on the NHR dataset \cite{nhr}. }
\label{tbl_nhr}
\end{table}

\noindent
\textbf{Free-viewpoint Rendering.}
We compare AT-GS with state-of-the-art methods for novel view synthesis in dynamic scenes, following the official implementations of these methods. These methods are categorized into two groups: (1) holistic approaches, such as 4K4D \cite{xu20244k4d} and SpacetimeGaussians (STG) \cite{stg}, which optimize entire video sequences as a whole; and (2) incremental methods, including 3DGStream \cite{sun20243dgstream} and NeuS2 \cite{wang2023neus2}, which train each frame sequentially.

Qualitative and quantitative comparisons on the DNA-Rendering dataset \cite{2023dnarendering} are presented in \cref{fig_dna_compare} and \cref{tbl_dna}, respectively. To quantitatively evaluate the rendering quality, we report the Peak Signal-to-Noise Ratio (PSNR), Structural Similarity Index (SSIM), and Learned Perceptual Image Patch Similarity (LPIPS) \cite{lpips} based on the VGG network \cite{vgg}. These metrics are calculated and averaged across all testing views and frames. As shown in \cref{tbl_dna}, our method outperforms existing approaches across all three metrics. Specifically, STG and 3DGStream struggle to recover areas with fast motion, such as the hands, as illustrated in both scenes of \cref{fig_dna_compare}. Additionally, NeuS2 fails to reconstruct fine-grained details like the basket frame in the second row of \cref{fig_dna_compare}. 4K4D, despite achieving photo-realistic appearances in dynamic regions, suffers from artifacts around object boundaries and struggles with non-Lambertian surfaces, such as the camera lens. In contrast, our method synthesizes novel views with higher visual fidelity, even in complex dynamic scenes.

We further evaluate our method on the NHR dataset \cite{nhr}. As demonstrated in \cref{tbl_nhr} and \cref{fig_nhr_compare}, our method achieves rendering quality comparable to 4K4D \cite{xu20244k4d} and NeuS2 \cite{wang2023neus2}, while significantly outperforming other methods. Notably, training 4K4D 
takes more than a day, whereas our method requires only about 1.5 hours. 

\begin{figure}[t]
\setlength{\abovecaptionskip}{2pt} 
\begin{center}
\includegraphics[trim={0cm 0.cm 0.0cm 0.cm},clip,width=0.999\linewidth]{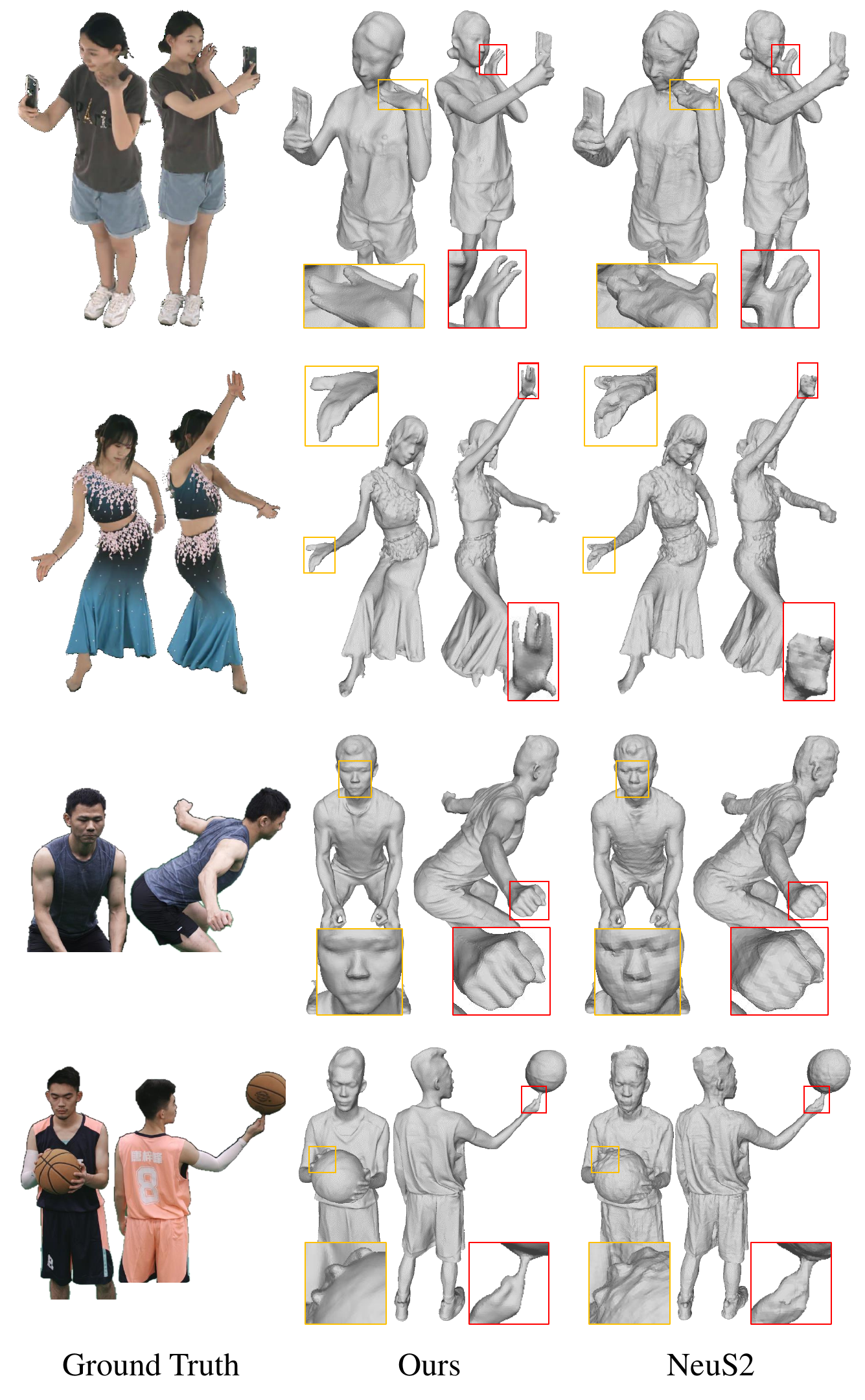}
\end{center}
  \caption{Qualitative comparison of surface reconstruction on the DNA-Rendering and NHR datasets. Compared to the state-of-the-art method, our approach produces superior results with finer and more accurate geometric details.}
\label{fig_mesh_compare}
\end{figure}

\noindent
\textbf{Surface reconstruction.}
We compare our method with NeuS2 \cite{wang2023neus2}, a state-of-the-art neural scene reconstruction method, to evaluate the geometric quality of dynamic reconstructions. Since ground truth geometry is not available for either dataset, we provide a qualitative comparison in \cref{fig_mesh_compare}. Specifically, NeuS2 tends to produce blurry surfaces on fine details, such as fingers, and is prone to noisy artifacts. In contrast, our method achieves more accurate geometry reconstruction with enhanced detail, even for thin objects like the phone in the first row and in challenging areas like the occluded upper body in the third row.

\subsection{Ablation Study} \label{sec:abl}
\begin{figure}[t]
\begin{center}
\includegraphics[trim={0cm 0.4cm 0.0cm 0.cm},clip,width=0.999\linewidth]{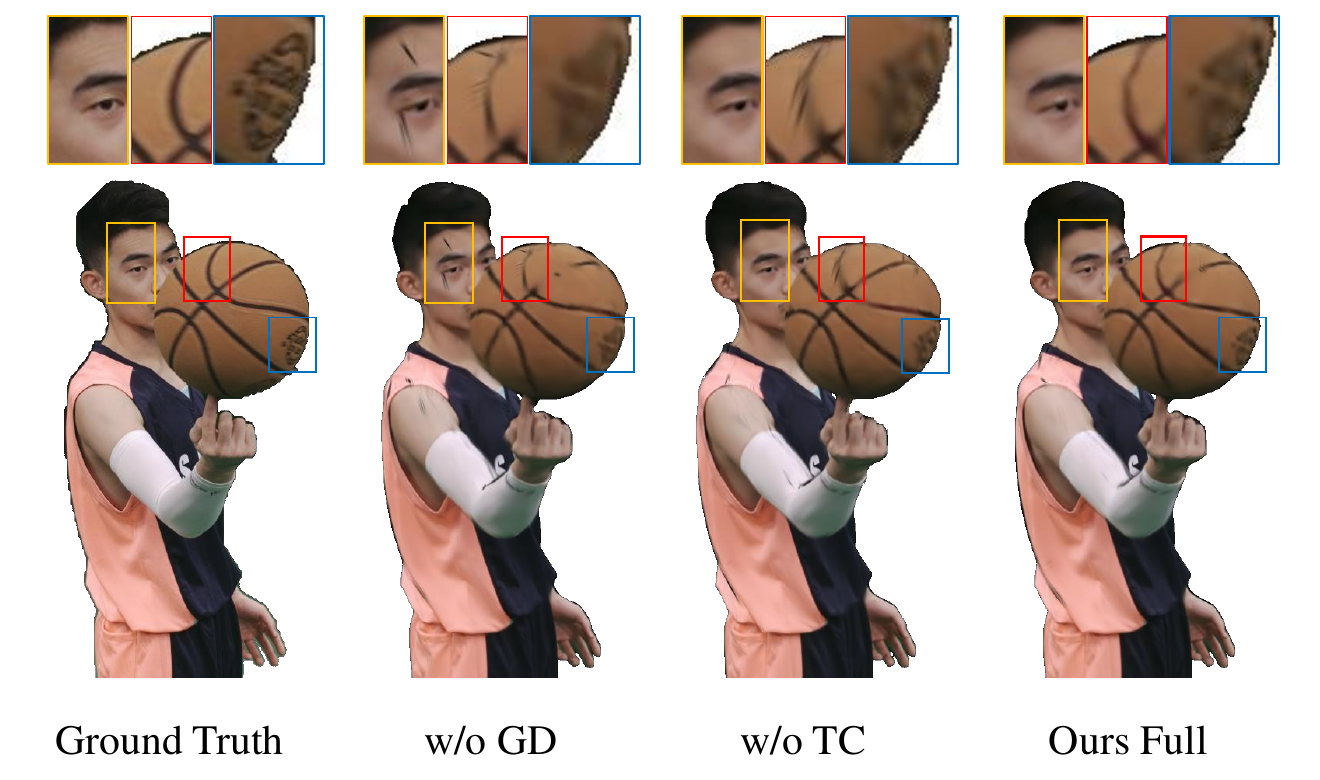}
\end{center}
  \caption{  Qualitative ablation study on the NHR dataset.  }
\label{fig_abl}
\end{figure}

To evaluate the contribution of our proposed components, we conduct both quantitative and qualitative analyses across all sequences from the NHR dataset. The “w/o GD” variant replaces the proposed gradient-aware densification strategy with conventional densification. As shown in \cref{tbl_abl} and \cref{fig_abl}, this modification introduces visual artifacts and results in inferior rendering, particularly in terms of perceptual quality (LPIPS). Similarly, the “w/o TC” variant omits the proposed curvature-based temporal consistency loss, leading to degradation in the fidelity of view synthesis. Additionally, an ablation comparison of this variant on dynamic meshes is provided in the supplementary video, highlighting the effectiveness of temporal consistency in maintaining visual coherence over time.

\begin{table} 
\centering
\scalebox{0.9}{ 
\begin{tabular}{lccc}
\toprule
Method & PSNR$\uparrow$ & SSIM$\uparrow$ & LPIPS$\downarrow$ \\
\midrule
w/o GD + w/o TC & 33.468 & 0.9729 & 0.0549 \\
w/o GD & \cellcolor{yellow!35}33.505 & 0.9731 & 0.0545 \\
w/o TC & 33.499 & \cellcolor{yellow!35}0.9732 & \cellcolor{yellow!35}0.0541 \\
Ours Full & \cellcolor{red!35}33.547 & \cellcolor{red!35}0.9733 & \cellcolor{red!35}0.0539 \\
\bottomrule
\end{tabular}
}
\caption{Ablation study on the NHR dataset. Both Gradient-aware Densification (GD) and curvature-based Temporal Consistency (TC) contribute to improved overall results. Metrics are averaged across all scenes.}
\label{tbl_abl}
\end{table}

\section{Conclusion}
In this work, we present AT-GS, a novel approach for efficient dynamic surface reconstruction from multi-view videos. By introducing a unified and gradient-aware densification strategy, we optimize dynamic 3D Gaussians with fine-grained details, overcoming the challenges of local minima and ensuring high-fidelity reconstruction. Additionally, our temporal consistency approach, which enforces curvature map consistency across frames, addresses the issue of temporal jittering, leading to stable and coherent surface reconstructions. Through extensive experiments on diverse datasets, we demonstrate that our method outperforms existing approaches in terms of rendering quality and geometric accuracy. 


\noindent
\textbf{Limitations.} 
First, because we focus on fast dynamic reconstruction, the limited number of training iterations per frame may hinder performance in handling extremely challenging objects, such as the small sparkling sequins on skirts, as seen in \cref{fig_dna_compare}. 
Additionally, since the Gaussian representation for each frame is stored separately, the storage overhead scales linearly with the video length, reducing storage efficiency for very long sequences.
Addressing these limitations will be an avenue for future work.

\section*{Acknowledgements}
This work has partly been funded by the German \mbox{Federal} Ministry for Economic Affairs and Climate Action (\mbox{ToHyVe}, grant no. 01MT22002A).


\twocolumn[{
\begin{center}
    {\Large \textbf{Adaptive and Temporally Consistent Gaussian Surfels for Multi-view Dynamic Reconstruction \\ --Supplementary Material--}} \\[1em]
\end{center}
}
]




\section{Implementation Details} \label{sec:implem}
In all our experiments, training is conducted on a GPU server equipped with an AMD EPYC 9654 CPU and an NVIDIA RTX 6000 Ada GPU, utilizing the Adam optimizer \cite{adam}, PyTorch 2.3.1 \cite{PyTorch}, and CUDA 11.8. For each dynamic scene, we begin with static reconstruction using Gaussian surfels \cite{gsurfels} for the first frame, obtaining a surfel-based Gaussian representation from a sparse point cloud generated by COLMAP \cite{colmap}. For each subsequent frame, we initialize the scene from the previous frame and apply our coarse-to-fine training approach, with 200 iterations for the coarse stage and 800 iterations for the fine stage. Training takes 31.7 seconds per frame on the NHR dataset \cite{nhr} and 37.5 seconds per frame on the DNA-Rendering dataset \cite{2023dnarendering}.

In the coarse stage, the learning rate for the Neural Transformation Cache is set to 0.002. In the fine stage, our unified, adaptive densification of Gaussians starts at iteration 230 and ends at iteration 600, with a densification interval of 30 iterations. Additionally, the Gaussian opacity reset interval is set to 200 iterations. We set the spherical harmonics degree to 1 for the NHR dataset and 2 for the DNA-Rendering dataset, as the latter contains more non-Lambertian objects. All other hyperparameters are kept consistent with 3DGS \cite{3dgs}.

For the loss function, we set $\lambda_{o}$ to 0.01 and $\lambda_{m}$ to 0.1. Additionally, we gradually increase $\lambda_{m}$ from 0.01 to 0.11, while linearly decaying $\lambda_{t}$ from 0.04 to 0.02.


\section{Additional Dataset Details}
For the DNA-Rendering dataset \cite{2023dnarendering}, we evaluate our method on five widely used sequences: 0008\_01, 0012\_11, 0013\_01, 0013\_03, and 0013\_09, with images downsampled by a factor of 2 and cropped to focus on the foreground region. Following 4K4D \cite{xu20244k4d}, we select views 11, 25, 37, and 57 as testing views, with the remaining views used for training.
For all scenes in the NHR dataset \cite{nhr}, we reserve views 18, 28, 37, and 46 for evaluation, while the rest serve as the training set.

\section{Additional Ablation Study}
In this section, we quantitatively evaluate the effectiveness of our method in enhancing temporal consistency.  Specifically, we render dynamic mesh sequences from a fixed testing view and calculate SSIM, PSNR, and LPIPS between consecutive frames. Temporal consistency is then measured by averaging these metrics across the entire sequence, with higher scores indicating greater similarity between consecutive frames. Since the scene movement remains consistent for the same rendering view, more similar images across frames suggest higher temporal consistency. As shown in \cref{tbl_abl_addi}, our curvature-based temporal consistency (TC) module significantly improves smoothness across frames. Additionally, a qualitative evaluation of temporal consistency is provided in the supplementary video.

\begin{table}[h]
\centering
\scalebox{0.9}{ 
\begin{tabular}{lccc}
\toprule
Method & PSNR$\uparrow$ & SSIM$\uparrow$ & LPIPS$\downarrow$ \\
\midrule
w/o GD + w/o TC & 29.268 & 0.946 & 0.0145 \\
w/o GD & \cellcolor{yellow!35}29.569 & \cellcolor{yellow!35}0.9507 & \cellcolor{yellow!35}0.0129 \\
w/o TC & 29.271 & 0.9469 & 0.0145 \\
Ours Full & \cellcolor{red!35}29.589 & \cellcolor{red!35}0.9514 & \cellcolor{red!35}0.0129 \\
\bottomrule
\end{tabular}
}
\caption{Ablation study on the temporal consistency of rendered mesh videos on the NHR dataset.}
\label{tbl_abl_addi}
\end{table}

\section{More Results}

\begin{figure*}
\setlength{\abovecaptionskip}{2pt} 
\begin{center}
\includegraphics[trim={0cm 0.cm 0cm 0cm},clip,width=0.9\linewidth]{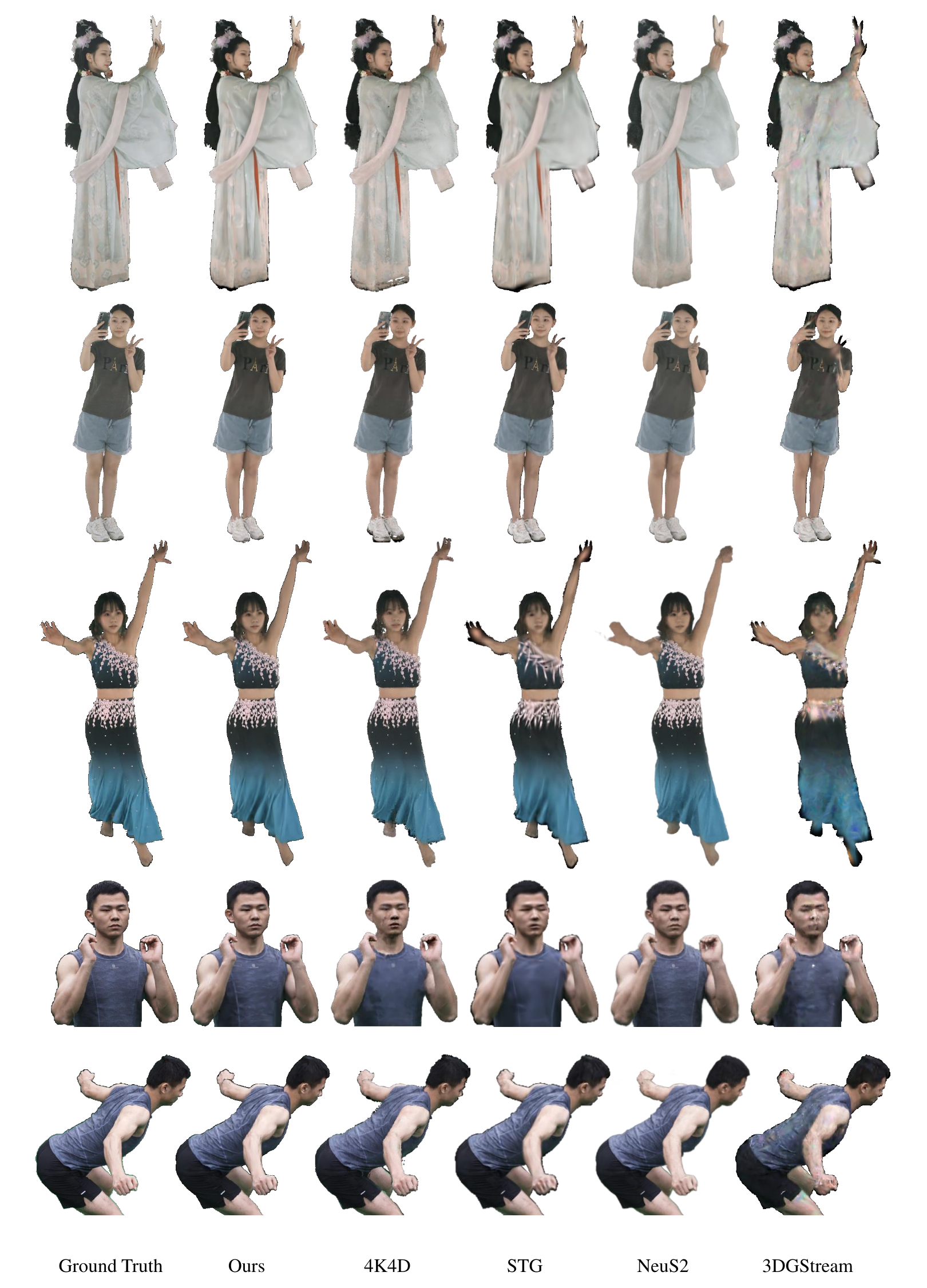}
\end{center}
  \caption{Additional qualitative comparison of novel view synthesis on the DNA-Rendering and NHR datasets.}
\label{fig_addi_compare}
\end{figure*}

\noindent \textbf{Free-Viewpoint Rendering.}
In \cref{tbl_dna_full} and \cref{tbl_full_nhr}, we provide a detailed per-scene quantitative comparison of our rendering results against various baselines on both the DNA-Rendering and NHR datasets. Additionally, as shown in \cref{fig_addi_compare}, our method consistently achieves photo-realistic rendering with fine-grained details.

\begin{figure*}
\setlength{\abovecaptionskip}{2pt} 
\begin{center}
\includegraphics[trim={0cm 0.cm 0.0cm 0.cm},clip,width=0.615\linewidth]{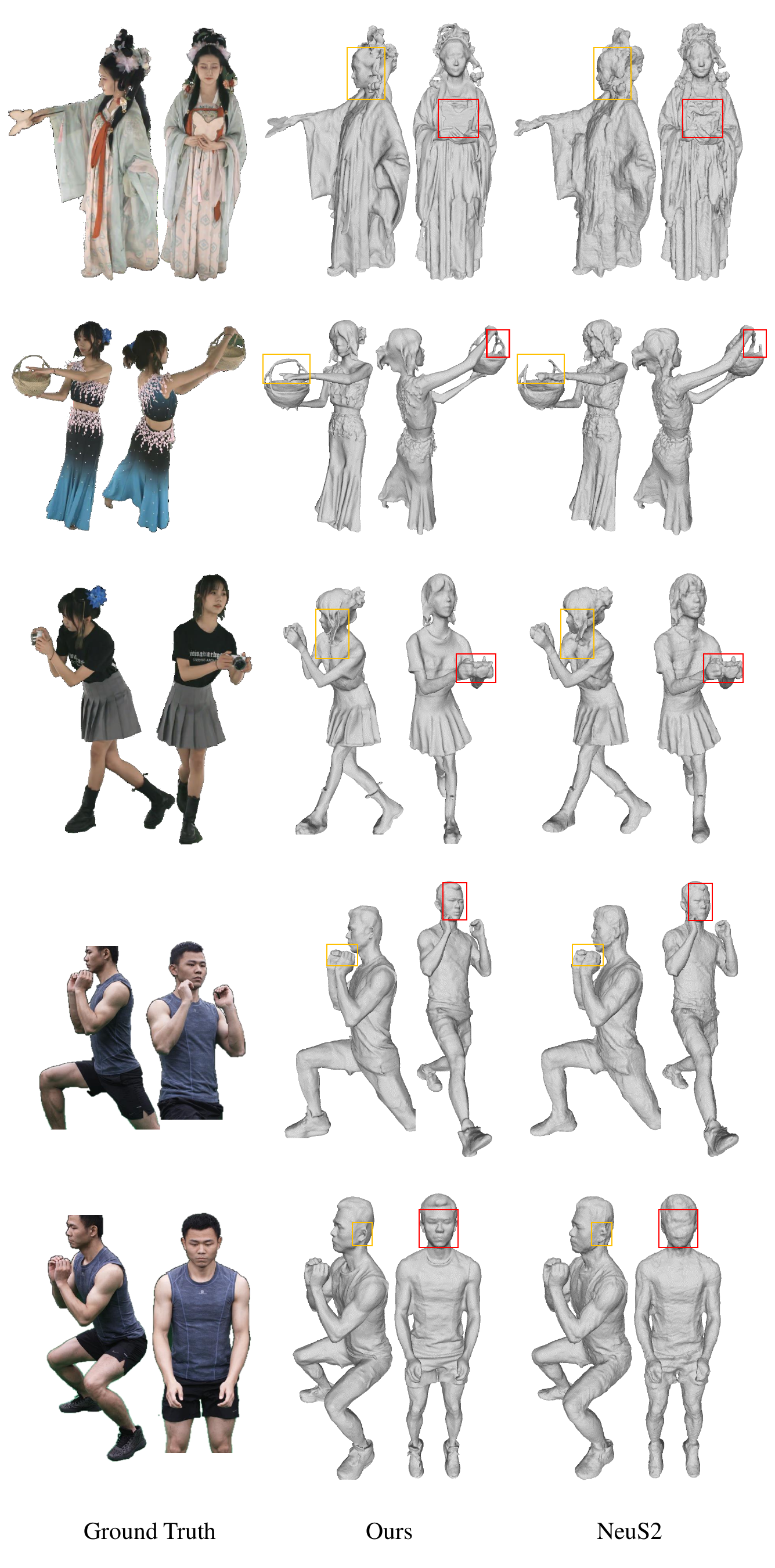}
\end{center}
  \caption{Additional comparison of surface reconstruction on the DNA-Rendering and NHR datasets.}
\label{fig_mesh_compare_addi}
\end{figure*}

\noindent \textbf{Surface Reconstruction.}
We include further qualitative comparisons of dynamic surface geometry on the DNA-Rendering and NHR datasets in \cref{fig_mesh_compare_addi}. Our method reconstructs high-quality surface meshes across various complex dynamic scenes.

\begin{table*}
\centering
\scalebox{0.9}{ 
\begin{tabular}{llccccccccc}
\toprule 
\multirow{2}{*}{Type} & \multirow{2}{*}{Method} & \multicolumn{3}{c}{0008\_01} & \multicolumn{3}{c}{0012\_11} & \multicolumn{3}{c}{0013\_01} \\ 
\cmidrule(lr){3-5} \cmidrule(lr){6-8} \cmidrule(lr){9-11}
 &  & PSNR$\uparrow$ & SSIM$\uparrow$ & LPIPS$\downarrow$ & PSNR$\uparrow$ & SSIM$\uparrow$ & LPIPS$\downarrow$ & PSNR$\uparrow$ & SSIM$\uparrow$ & LPIPS$\downarrow$ \\ 
\midrule
\multirow{2}{*}{Holistic} & 4K4D & \cellcolor{yellow!35}31.36 & \cellcolor{yellow!35}0.974 & \cellcolor{yellow!35}0.047 & \cellcolor{yellow!35}35.81 & \cellcolor{yellow!35}0.990 & \cellcolor{red!35}0.018 & \cellcolor{yellow!35}34.52 & \cellcolor{yellow!35}0.987 & \cellcolor{red!35}0.021 \\ 
 & STG & 24.08 & 0.944 & 0.068 & 33.55 & 0.986 & \cellcolor{yellow!35}0.023 & 25.47 & 0.957 & 0.047 \\ 
\midrule
\multirow{3}{*}{Incremental} & NeuS2 & 30.24 & \cellcolor{red!35}0.980 & 0.054 & 35.54 & \cellcolor{red!35}0.992 & \cellcolor{yellow!35}0.023 & 33.33 & \cellcolor{yellow!35}0.987 & 0.030 \\ 
 & 3DGStream & 27.46 & 0.960 & 0.075 & 33.88 & 0.986 & 0.033 & 29.14 & 0.969 & 0.047 \\ 
 & Ours & \cellcolor{red!35}32.07 & \cellcolor{red!35}0.980 & \cellcolor{red!35}0.039 & \cellcolor{red!35}37.03 & \cellcolor{red!35}0.992 & \cellcolor{red!35}0.018 & \cellcolor{red!35}35.46 & \cellcolor{red!35}0.988 & \cellcolor{yellow!35}0.022 \\ 
\bottomrule 
\toprule 
\multirow{2}{*}{Type} & \multirow{2}{*}{Method} & \multicolumn{3}{c}{0013\_03} & \multicolumn{3}{c}{0013\_09} & \multicolumn{3}{c}{\textbf{Average}} \\ 
\cmidrule(lr){3-5} \cmidrule(lr){6-8} \cmidrule(lr){9-11}
 &  & PSNR$\uparrow$ & SSIM$\uparrow$ & LPIPS$\downarrow$ & PSNR$\uparrow$ & SSIM$\uparrow$ & LPIPS$\downarrow$ & PSNR$\uparrow$ & SSIM$\uparrow$ & LPIPS$\downarrow$ \\ 
\midrule
\multirow{2}{*}{Holistic} & 4K4D & \cellcolor{yellow!35}34.41 & 0.986 & \cellcolor{yellow!35}0.022 & \cellcolor{yellow!35}36.48 & \cellcolor{yellow!35}0.989 & 0\cellcolor{red!35}.020 & \cellcolor{yellow!35}34.52 & 0.985 & \cellcolor{yellow!35}0.025 \\ 
 & STG & 27.49 & 0.965 & 0.037 & 31.84 & 0.977 & 0.031 & 28.49 & 0.966 & 0.041 \\ 
\midrule
\multirow{3}{*}{Incremental} & NeuS2 & 33.60 & \cellcolor{yellow!35}0.987 & 0.029 & 36.27 & \cellcolor{red!35}0.990 & \cellcolor{yellow!35}0.025 & 33.80 & \cellcolor{yellow!35}0.987 & 0.032 \\ 
 & 3DGStream & 29.78 & 0.972 & 0.045 & 33.63 & 0.982 & 0.037 & 30.78 & 0.974 & 0.047 \\ 
 & Ours & \cellcolor{red!35}35.43 & \cellcolor{red!35}0.988 & \cellcolor{red!35}0.020 & \cellcolor{red!35}37.19 & \cellcolor{red!35}0.990 & \cellcolor{red!35}0.020 & \cellcolor{red!35}35.44 & \cellcolor{red!35}0.988 & \cellcolor{red!35}0.024 \\ 
\bottomrule 
\end{tabular}
}
\caption{Per-scene quantitative results on the DNA-Rendering dataset \cite{2023dnarendering}. The best values are highlighted in \colorbox{red!35}{\textbf{red}}, and the second-best values in \colorbox{yellow!35}{\textbf{yellow}}. Our method achieves the highest rendering quality compared to all other baselines.}
\label{tbl_dna_full}
\end{table*}

\begin{table*}
\centering
\scalebox{0.9}{ 
\setlength{\tabcolsep}{1.6pt}
\begin{tabular}{lccccccccccccccc}
\toprule 
\multirow{2}{*}{Method} & \multicolumn{3}{c}{sport\_1} & \multicolumn{3}{c}{sport\_2} & \multicolumn{3}{c}{sport\_3} & \multicolumn{3}{c}{basketball} & \multicolumn{3}{c}{\textbf{Average}} \\ 
\cmidrule(lr){2-4} \cmidrule(lr){5-7} \cmidrule(lr){8-10} \cmidrule(lr){11-13} \cmidrule(lr){14-16}
 & PSNR$\uparrow$ & SSIM$\uparrow$ & LPIPS$\downarrow$ & PSNR$\uparrow$ & SSIM$\uparrow$ & LPIPS$\downarrow$ & PSNR$\uparrow$ & SSIM$\uparrow$ & LPIPS$\downarrow$ & PSNR$\uparrow$ & SSIM$\uparrow$ & LPIPS$\downarrow$ & PSNR$\uparrow$ & SSIM$\uparrow$ & LPIPS$\downarrow$ \\ 
\midrule
 4K4D & 33.37 & \cellcolor{red!35}0.975 & \cellcolor{red!35}0.026 & \cellcolor{red!35}34.57 & 0.968 & \cellcolor{yellow!35}0.052 & \cellcolor{red!35}34.19 & 0.968 & \cellcolor{yellow!35}0.051 & \cellcolor{red!35}32.49 & \cellcolor{red!35}0.977 & \cellcolor{red!35}0.027 & \cellcolor{red!35}33.65 & \cellcolor{yellow!35}0.972 & \cellcolor{red!35}0.039 \\ 
 STG & 28.65 & 0.952 & 0.068 & 29.88 & 0.958 & 0.065 & 26.34 & 0.940 & 0.084 & 27.35 & 0.949 & 0.080 & 28.05 & 0.949 & 0.074 \\ 
\midrule
 NeuS2 & \cellcolor{yellow!35}33.53 & \cellcolor{red!35}0.975 & \cellcolor{yellow!35}0.038 & 33.62 & \cellcolor{yellow!35}0.971 & \cellcolor{red!35}0.047 & 33.35 & \cellcolor{yellow!35}0.972 & \cellcolor{red!35}0.044 & 31.66 & 0.970 & \cellcolor{yellow!35}0.057 & 33.04 & \cellcolor{yellow!35}0.972 & \cellcolor{yellow!35}0.047 \\ 
 3DGStream & 31.73 & 0.960 & 0.070 & 31.12 & 0.955 & 0.082 & 30.86 & 0.954 & 0.083 & 29.08 &  0.951 & 0.096 & 30.70 & 0.955 & 0.083 \\ 
 Ours & \cellcolor{red!35}33.64 & \cellcolor{yellow!35}0.974 & 0.046 & \cellcolor{yellow!35}34.42 & \cellcolor{red!35}0.973 & 0.056 & \cellcolor{yellow!35}34.14 & \cellcolor{red!35}0.974 & 0.052 & \cellcolor{yellow!35}31.99 & \cellcolor{yellow!35}0.972 & 0.060 & \cellcolor{yellow!35}33.55 & \cellcolor{red!35}0.973 & 0.054 \\ 
\bottomrule 
\end{tabular}
}
\caption{Per-scene quantitative results on the NHR dataset \cite{nhr}. }
\label{tbl_full_nhr}
\end{table*}

\section{Supplementary Video}
The supplementary video includes the following:
\begin{itemize}
    \item Additional ablation study on the impact of temporal consistency loss on dynamic surface meshes.
    \item A comparison between our method and NeuS2 \cite{wang2023neus2} on dynamic surface meshes.
    \item Additional results showcasing free-viewpoint renderings of both color images and surface meshes.
\end{itemize}

\section{Potential Societal Impact}
While AT-GS advances dynamic surface reconstruction and novel view synthesis, its deployment carries potential negative societal impacts. When combined with generative technology, it could be misused to create hyper-realistic deepfakes or synthetic media, leading to disinformation, privacy breaches, and security risks. The high-fidelity reconstruction capabilities may also be exploited for intrusive surveillance, further raising privacy concerns. Additionally, although more efficient than some methods, the computational demands of AT-GS could contribute to environmental impact due to energy consumption, especially at scale. It is essential for researchers to remain vigilant and prioritize ethical use, alongside exploring safeguards to mitigate these risks.

{\small
\bibliographystyle{ieee_fullname}
\bibliography{egbib}

\begin{thebibliography}{10}\itemsep=-1pt

\bibitem{E-D3DGS}
Jeongmin Bae, Seoha Kim, Youngsik Yun, Hahyun Lee, Gun Bang, and Youngjung Uh.
\newblock Per-gaussian embedding-based deformation for deformable 3d gaussian splatting.
\newblock {\em arXiv preprint arXiv:2404.03613}, 2024.

\bibitem{ndr}
Hongrui Cai, Wanquan Feng, Xuetao Feng, Yan Wang, and Juyong Zhang.
\newblock Neural surface reconstruction of dynamic scenes with monocular rgb-d camera.
\newblock {\em Advances in Neural Information Processing Systems}, 35:967--981, 2022.

\bibitem{cao2023hexplane}
Ang Cao and Justin Johnson.
\newblock Hexplane: A fast representation for dynamic scenes.
\newblock In {\em Proceedings of the IEEE/CVF Conference on Computer Vision and Pattern Recognition}, pages 130--141, 2023.

\bibitem{chen2022tensorf}
Anpei Chen, Zexiang Xu, Andreas Geiger, Jingyi Yu, and Hao Su.
\newblock Tensorf: Tensorial radiance fields.
\newblock In {\em European conference on computer vision}, pages 333--350. Springer, 2022.

\bibitem{chen2024pgsr}
Danpeng Chen, Hai Li, Weicai Ye, Yifan Wang, Weijian Xie, Shangjin Zhai, Nan Wang, Haomin Liu, Hujun Bao, and Guofeng Zhang.
\newblock Pgsr: Planar-based gaussian splatting for efficient and high-fidelity surface reconstruction.
\newblock {\em arXiv preprint arXiv:2406.06521}, 2024.

\bibitem{chen2023dynamic}
Decai Chen, Haofei Lu, Ingo Feldmann, Oliver Schreer, and Peter Eisert.
\newblock Dynamic multi-view scene reconstruction using neural implicit surface.
\newblock In {\em ICASSP 2023-2023 IEEE International Conference on Acoustics, Speech and Signal Processing (ICASSP)}, pages 1--5. IEEE, 2023.

\bibitem{chen2021accurate}
Decai Chen, Markus Worchel, Ingo Feldmann, Oliver Schreer, and Peter Eisert.
\newblock Accurate human body reconstruction for volumetric video.
\newblock In {\em 2021 International Conference on 3D Immersion (IC3D)}, pages 1--8. IEEE, 2021.

\bibitem{dneus}
Decai Chen, Peng Zhang, Ingo Feldmann, Oliver Schreer, and Peter Eisert.
\newblock Recovering fine details for neural implicit surface reconstruction.
\newblock In {\em Proceedings of the IEEE/CVF Winter Conference on Applications of Computer Vision}, pages 4330--4339, 2023.

\bibitem{chen2023neusg}
Hanlin Chen, Chen Li, and Gim~Hee Lee.
\newblock Neusg: Neural implicit surface reconstruction with 3d gaussian splatting guidance.
\newblock {\em arXiv preprint arXiv:2312.00846}, 2023.

\bibitem{2023dnarendering}
Wei Cheng, Ruixiang Chen, Wanqi Yin, Siming Fan, Keyu Chen, Honglin He, Huiwen Luo, Zhongang Cai, Jingbo Wang, Yang Gao, Zhengming Yu, Zhengyu Lin, Daxuan Ren, Lei Yang, Ziwei Liu, Chen~Change Loy, Chen Qian, Wayne Wu, Dahua Lin, Bo Dai, and Kwan-Yee Lin.
\newblock Dna-rendering: A diverse neural actor repository for high-fidelity human-centric rendering.
\newblock {\em arXiv preprint}, arXiv:2307.10173, 2023.

\bibitem{cao2024mvsformerpp}
Xinlin~Ren Chenjie~Cao and Yanwei Fu.
\newblock Mvsformer++: Revealing the devil in transformer's details for multi-view stereo.
\newblock In {\em International Conference on Learning Representations (ICLR)}, 2024.

\bibitem{choe2023spacetime}
Jaesung Choe, Christopher Choy, Jaesik Park, In~So Kweon, and Anima Anandkumar.
\newblock Spacetime surface regularization for neural dynamic scene reconstruction.
\newblock In {\em Proceedings of the IEEE/CVF International Conference on Computer Vision}, pages 17871--17881, 2023.

\bibitem{gsurfels}
Pinxuan Dai, Jiamin Xu, Wenxiang Xie, Xinguo Liu, Huamin Wang, and Weiwei Xu.
\newblock High-quality surface reconstruction using gaussian surfels.
\newblock In {\em ACM SIGGRAPH 2024 Conference Papers}, pages 1--11, 2024.

\bibitem{das2024neural}
Devikalyan Das, Christopher Wewer, Raza Yunus, Eddy Ilg, and Jan~Eric Lenssen.
\newblock Neural parametric gaussians for monocular non-rigid object reconstruction.
\newblock In {\em Proceedings of the IEEE/CVF Conference on Computer Vision and Pattern Recognition}, pages 10715--10725, 2024.

\bibitem{ding2022transmvsnet}
Yikang Ding, Wentao Yuan, Qingtian Zhu, Haotian Zhang, Xiangyue Liu, Yuanjiang Wang, and Xiao Liu.
\newblock Transmvsnet: Global context-aware multi-view stereo network with transformers.
\newblock In {\em Proceedings of the IEEE/CVF conference on computer vision and pattern recognition}, pages 8585--8594, 2022.

\bibitem{duan20244drotor}
Yuanxing Duan, Fangyin Wei, Qiyu Dai, Yuhang He, Wenzheng Chen, and Baoquan Chen.
\newblock 4d-rotor gaussian splatting: Towards efficient novel view synthesis for dynamic scenes.
\newblock In {\em ACM SIGGRAPH 2024 Conference Papers}, pages 1--11, 2024.

\bibitem{Kplanes}
Sara Fridovich-Keil, Giacomo Meanti, Frederik~Rahb{\ae}k Warburg, Benjamin Recht, and Angjoo Kanazawa.
\newblock K-planes: Explicit radiance fields in space, time, and appearance.
\newblock In {\em Proceedings of the IEEE/CVF Conference on Computer Vision and Pattern Recognition}, pages 12479--12488, 2023.

\bibitem{Fu2022GeoNeus}
Qiancheng Fu, Qingshan Xu, Yew-Soon Ong, and Wenbing Tao.
\newblock Geo-neus: Geometry-consistent neural implicit surfaces learning for multi-view reconstruction.
\newblock {\em Advances in Neural Information Processing Systems (NeurIPS)}, 2022.

\bibitem{DynamicNeRF}
Chen Gao, Ayush Saraf, Johannes Kopf, and Jia-Bin Huang.
\newblock Dynamic view synthesis from dynamic monocular video.
\newblock In {\em Proceedings of the IEEE/CVF International Conference on Computer Vision}, pages 5712--5721, 2021.

\bibitem{gao2024gaussianflow}
Quankai Gao, Qiangeng Xu, Zhe Cao, Ben Mildenhall, Wenchao Ma, Le Chen, Danhang Tang, and Ulrich Neumann.
\newblock Gaussianflow: Splatting gaussian dynamics for 4d content creation.
\newblock {\em arXiv preprint arXiv:2403.12365}, 2024.

\bibitem{gaussian_splatting_issue_89}
graphdeco inria.
\newblock Issue \#89: Question and answers about randomness in the optimization progress of the gaussian splatting.
\newblock \url{https://github.com/graphdeco-inria/gaussian-splatting/issues/89}, 2024.

\bibitem{guedon2024sugar}
Antoine Gu{\'e}don and Vincent Lepetit.
\newblock Sugar: Surface-aligned gaussian splatting for efficient 3d mesh reconstruction and high-quality mesh rendering.
\newblock In {\em Proceedings of the IEEE/CVF Conference on Computer Vision and Pattern Recognition}, pages 5354--5363, 2024.

\bibitem{guo2024motion}
Zhiyang Guo, Wengang Zhou, Li Li, Min Wang, and Houqiang Li.
\newblock Motion-aware 3d gaussian splatting for efficient dynamic scene reconstruction.
\newblock {\em arXiv preprint arXiv:2403.11447}, 2024.

\bibitem{2dgs}
Binbin Huang, Zehao Yu, Anpei Chen, Andreas Geiger, and Shenghua Gao.
\newblock 2d gaussian splatting for geometrically accurate radiance fields.
\newblock In {\em ACM SIGGRAPH 2024 Conference Papers}, pages 1--11, 2024.

\bibitem{huang2024scgs}
Yi-Hua Huang, Yang-Tian Sun, Ziyi Yang, Xiaoyang Lyu, Yan-Pei Cao, and Xiaojuan Qi.
\newblock Sc-gs: Sparse-controlled gaussian splatting for editable dynamic scenes.
\newblock In {\em Proceedings of the IEEE/CVF Conference on Computer Vision and Pattern Recognition}, pages 4220--4230, 2024.

\bibitem{icsik2023humanrf}
Mustafa I{\c{s}}{\i}k, Martin R{\"u}nz, Markos Georgopoulos, Taras Khakhulin, Jonathan Starck, Lourdes Agapito, and Matthias Nie{\ss}ner.
\newblock Humanrf: High-fidelity neural radiance fields for humans in motion.
\newblock {\em ACM Transactions on Graphics (TOG)}, 42(4):1--12, 2023.

\bibitem{jiang2024hifi4g}
Yuheng Jiang, Zhehao Shen, Penghao Wang, Zhuo Su, Yu Hong, Yingliang Zhang, Jingyi Yu, and Lan Xu.
\newblock Hifi4g: High-fidelity human performance rendering via compact gaussian splatting.
\newblock In {\em Proceedings of the IEEE/CVF Conference on Computer Vision and Pattern Recognition}, pages 19734--19745, 2024.

\bibitem{johnson2023unbiased}
Erik Johnson, Marc Habermann, Soshi Shimada, Vladislav Golyanik, and Christian Theobalt.
\newblock Unbiased 4d: Monocular 4d reconstruction with a neural deformation model.
\newblock In {\em Proceedings of the IEEE/CVF Conference on Computer Vision and Pattern Recognition}, pages 6598--6607, 2023.

\bibitem{katsumata2023efficient}
Kai Katsumata, Duc~Minh Vo, and Hideki Nakayama.
\newblock An efficient 3d gaussian representation for monocular/multi-view dynamic scenes.
\newblock {\em arXiv preprint arXiv:2311.12897}, 2023.

\bibitem{kazhdan2013screened}
Michael Kazhdan and Hugues Hoppe.
\newblock Screened poisson surface reconstruction.
\newblock {\em ACM Transactions on Graphics (ToG)}, 32(3):1--13, 2013.

\bibitem{3dgs}
Bernhard Kerbl, Georgios Kopanas, Thomas Leimk{\"u}hler, and George Drettakis.
\newblock 3d gaussian splatting for real-time radiance field rendering.
\newblock {\em ACM Trans. Graph.}, 42(4):139--1, 2023.

\bibitem{adam}
Diederik Kingma and Jimmy Ba.
\newblock Adam: A method for stochastic optimization.
\newblock {\em International Conference on Learning Representations}, 12 2014.

\bibitem{kirschstein2023nersemble}
Tobias Kirschstein, Shenhan Qian, Simon Giebenhain, Tim Walter, and Matthias Nie{\ss}ner.
\newblock Nersemble: Multi-view radiance field reconstruction of human heads.
\newblock {\em ACM Transactions on Graphics (TOG)}, 42(4):1--14, 2023.

\bibitem{kmh24}
P. Knoll, W. Morgenstern, A. Hilsmann, and P. Eisert.
\newblock Animating nerfs from texture space: A framework for pose-dependent rendering of human performances.
\newblock In {\em Proc. Int. Conf. on Computer Vision Theory and Applications (VISAPP)}, pages 404--413, Rome, Italy, 2024.

\bibitem{kratimenos2023dynmf}
Agelos Kratimenos, Jiahui Lei, and Kostas Daniilidis.
\newblock Dynmf: Neural motion factorization for real-time dynamic view synthesis with 3d gaussian splatting.
\newblock {\em arXiv preprint arXiv:2312.00112}, 2023.

\bibitem{li2022streamrf}
Lingzhi Li, Zhen Shen, Zhongshu Wang, Li Shen, and Ping Tan.
\newblock Streaming radiance fields for 3d video synthesis.
\newblock {\em Advances in Neural Information Processing Systems}, 35:13485--13498, 2022.

\bibitem{DyNeRF}
Tianye Li, Mira Slavcheva, Michael Zollhoefer, Simon Green, Christoph Lassner, Changil Kim, Tanner Schmidt, Steven Lovegrove, Michael Goesele, Richard Newcombe, et~al.
\newblock Neural 3d video synthesis from multi-view video.
\newblock In {\em Proceedings of the IEEE/CVF Conference on Computer Vision and Pattern Recognition}, pages 5521--5531, 2022.

\bibitem{stg}
Zhan Li, Zhang Chen, Zhong Li, and Yi Xu.
\newblock Spacetime gaussian feature splatting for real-time dynamic view synthesis.
\newblock In {\em Proceedings of the IEEE/CVF Conference on Computer Vision and Pattern Recognition}, pages 8508--8520, 2024.

\bibitem{li2023neuralangelo}
Zhaoshuo Li, Thomas M{\"u}ller, Alex Evans, Russell~H Taylor, Mathias Unberath, Ming-Yu Liu, and Chen-Hsuan Lin.
\newblock Neuralangelo: High-fidelity neural surface reconstruction.
\newblock In {\em Proceedings of the IEEE/CVF Conference on Computer Vision and Pattern Recognition}, pages 8456--8465, 2023.

\bibitem{nsff}
Zhengqi Li, Simon Niklaus, Noah Snavely, and Oliver Wang.
\newblock Neural scene flow fields for space-time view synthesis of dynamic scenes.
\newblock In {\em Proceedings of the IEEE/CVF Conference on Computer Vision and Pattern Recognition (CVPR)}, 2021.

\bibitem{liang2023gaufre}
Yiqing Liang, Numair Khan, Zhengqin Li, Thu Nguyen-Phuoc, Douglas Lanman, James Tompkin, and Lei Xiao.
\newblock Gaufre: Gaussian deformation fields for real-time dynamic novel view synthesis.
\newblock {\em arXiv preprint arXiv:2312.11458}, 2023.

\bibitem{Im4D}
Haotong Lin, Sida Peng, Zhen Xu, Tao Xie, Xingyi He, Hujun Bao, and Xiaowei Zhou.
\newblock High-fidelity and real-time novel view synthesis for dynamic scenes.
\newblock In {\em SIGGRAPH Asia 2023 Conference Papers}, pages 1--9, 2023.

\bibitem{lin2024gaussian}
Youtian Lin, Zuozhuo Dai, Siyu Zhu, and Yao Yao.
\newblock Gaussian-flow: 4d reconstruction with dynamic 3d gaussian particle.
\newblock In {\em Proceedings of the IEEE/CVF Conference on Computer Vision and Pattern Recognition}, pages 21136--21145, 2024.

\bibitem{DGMesh}
Isabella Liu, Hao Su, and Xiaolong Wang.
\newblock Dynamic gaussians mesh: Consistent mesh reconstruction from monocular videos.
\newblock {\em arXiv preprint arXiv:2404.12379}, 2024.

\bibitem{luiten2024dynamic3dg}
Jonathon Luiten, Georgios Kopanas, Bastian Leibe, and Deva Ramanan.
\newblock Dynamic 3d gaussians: Tracking by persistent dynamic view synthesis.
\newblock In {\em 2024 International Conference on 3D Vision (3DV)}, pages 800--809. IEEE, 2024.

\bibitem{mags}
Shaojie Ma, Yawei Luo, and Yi Yang.
\newblock Reconstructing and simulating dynamic 3d objects with mesh-adsorbed gaussian splatting.
\newblock {\em arXiv preprint arXiv:2406.01593}, 2024.

\bibitem{nerf}
Ben Mildenhall, Pratul~P. Srinivasan, Matthew Tancik, Jonathan~T. Barron, Ravi Ramamoorthi, and Ren Ng.
\newblock Nerf: Representing scenes as neural radiance fields for view synthesis.
\newblock In {\em ECCV}, 2020.

\bibitem{muller2022instant}
Thomas M{\"u}ller, Alex Evans, Christoph Schied, and Alexander Keller.
\newblock Instant neural graphics primitives with a multiresolution hash encoding.
\newblock {\em arXiv preprint arXiv:2201.05989}, 2022.

\bibitem{caching}
Thomas M{\"u}ller, Fabrice Rousselle, Jan Nov{\'a}k, and Alexander Keller.
\newblock Real-time neural radiance caching for path tracing.
\newblock {\em arXiv preprint arXiv:2106.12372}, 2021.

\bibitem{nerfies}
Keunhong Park, Utkarsh Sinha, Jonathan~T. Barron, Sofien Bouaziz, Dan~B Goldman, Steven~M. Seitz, and Ricardo Martin-Brualla.
\newblock Nerfies: Deformable neural radiance fields.
\newblock {\em ICCV}, 2021.

\bibitem{hypernerf}
Keunhong Park, Utkarsh Sinha, Peter Hedman, Jonathan~T. Barron, Sofien Bouaziz, Dan~B Goldman, Ricardo Martin-Brualla, and Steven~M. Seitz.
\newblock Hypernerf: A higher-dimensional representation for topologically varying neural radiance fields.
\newblock {\em ACM Trans. Graph.}, 40(6), dec 2021.

\bibitem{PyTorch}
Adam Paszke, Sam Gross, Francisco Massa, Adam Lerer, James Bradbury, Gregory Chanan, Trevor Killeen, Zeming Lin, Natalia Gimelshein, Luca Antiga, Alban Desmaison, Andreas Kopf, Edward Yang, Zachary DeVito, Martin Raison, Alykhan Tejani, Sasank Chilamkurthy, Benoit Steiner, Lu Fang, Junjie Bai, and Soumith Chintala.
\newblock Pytorch: An imperative style, high-performance deep learning library.
\newblock In {\em Advances in Neural Information Processing Systems 32}, pages 8024--8035. Curran Associates, Inc., 2019.

\bibitem{d-nerf}
Albert Pumarola, Enric Corona, Gerard Pons-Moll, and Francesc Moreno-Noguer.
\newblock {D-NeRF: Neural Radiance Fields for Dynamic Scenes}.
\newblock In {\em Proceedings of the IEEE/CVF Conference on Computer Vision and Pattern Recognition}, 2020.

\bibitem{colmap}
Johannes~Lutz Sch\"{o}nberger and Jan-Michael Frahm.
\newblock {Structure-from-Motion Revisited}.
\newblock In {\em Conference on Computer Vision and Pattern Recognition (CVPR)}, 2016.

\bibitem{shao2023tensor4d}
Ruizhi Shao, Zerong Zheng, Hanzhang Tu, Boning Liu, Hongwen Zhang, and Yebin Liu.
\newblock Tensor4d: Efficient neural 4d decomposition for high-fidelity dynamic reconstruction and rendering.
\newblock In {\em Proceedings of the IEEE/CVF Conference on Computer Vision and Pattern Recognition}, pages 16632--16642, 2023.

\bibitem{shaw2023swags}
Richard Shaw, Jifei Song, Arthur Moreau, Michal Nazarczuk, Sibi Catley-Chandar, Helisa Dhamo, and Eduardo Perez-Pellitero.
\newblock Swags: Sampling windows adaptively for dynamic 3d gaussian splatting.
\newblock {\em arXiv preprint arXiv:2312.13308}, 2023.

\bibitem{vgg}
Karen Simonyan and Andrew Zisserman.
\newblock Very deep convolutional networks for large-scale image recognition.
\newblock {\em arXiv preprint arXiv:1409.1556}, 2014.

\bibitem{song2023nerfplayer}
Liangchen Song, Anpei Chen, Zhong Li, Zhang Chen, Lele Chen, Junsong Yuan, Yi Xu, and Andreas Geiger.
\newblock Nerfplayer: A streamable dynamic scene representation with decomposed neural radiance fields.
\newblock {\em IEEE Transactions on Visualization and Computer Graphics}, 29(5):2732--2742, 2023.

\bibitem{sun20243dgstream}
Jiakai Sun, Han Jiao, Guangyuan Li, Zhanjie Zhang, Lei Zhao, and Wei Xing.
\newblock 3dgstream: On-the-fly training of 3d gaussians for efficient streaming of photo-realistic free-viewpoint videos.
\newblock In {\em Proceedings of the IEEE/CVF Conference on Computer Vision and Pattern Recognition}, pages 20675--20685, 2024.

\bibitem{teed2020raft}
Zachary Teed and Jia Deng.
\newblock Raft: Recurrent all-pairs field transforms for optical flow.
\newblock In {\em Computer Vision--ECCV 2020: 16th European Conference, Glasgow, UK, August 23--28, 2020, Proceedings, Part II 16}, pages 402--419. Springer, 2020.

\bibitem{tian2023mononerf}
Fengrui Tian, Shaoyi Du, and Yueqi Duan.
\newblock Mononerf: Learning a generalizable dynamic radiance field from monocular videos.
\newblock In {\em Proceedings of the IEEE/CVF International Conference on Computer Vision}, pages 17903--17913, 2023.

\bibitem{NR-NeRF}
Edgar Tretschk, Ayush Tewari, Vladislav Golyanik, Michael Zollh\"{o}fer, Christoph Lassner, and Christian Theobalt.
\newblock Non-rigid neural radiance fields: Reconstruction and novel view synthesis of a dynamic scene from monocular video.
\newblock In {\em {IEEE} International Conference on Computer Vision ({ICCV})}. {IEEE}, 2021.

\bibitem{wan2024superpoint}
Diwen Wan, Ruijie Lu, and Gang Zeng.
\newblock Superpoint gaussian splatting for real-time high-fidelity dynamic scene reconstruction.
\newblock {\em arXiv preprint arXiv:2406.03697}, 2024.

\bibitem{wang2023mixed}
Feng Wang, Sinan Tan, Xinghang Li, Zeyue Tian, Yafei Song, and Huaping Liu.
\newblock Mixed neural voxels for fast multi-view video synthesis.
\newblock In {\em Proceedings of the IEEE/CVF International Conference on Computer Vision}, pages 19706--19716, 2023.

\bibitem{ReRF}
Liao Wang, Qiang Hu, Qihan He, Ziyu Wang, Jingyi Yu, Tinne Tuytelaars, Lan Xu, and Minye Wu.
\newblock Neural residual radiance fields for streamably free-viewpoint videos.
\newblock In {\em Proceedings of the IEEE/CVF Conference on Computer Vision and Pattern Recognition}, pages 76--87, 2023.

\bibitem{neus}
Peng Wang, Lingjie Liu, Yuan Liu, Christian Theobalt, Taku Komura, and Wenping Wang.
\newblock Neus: Learning neural implicit surfaces by volume rendering for multi-view reconstruction.
\newblock {\em NeurIPS}, 2021.

\bibitem{wang2023neus2}
Yiming Wang, Qin Han, Marc Habermann, Kostas Daniilidis, Christian Theobalt, and Lingjie Liu.
\newblock Neus2: Fast learning of neural implicit surfaces for multi-view reconstruction.
\newblock In {\em Proceedings of the IEEE/CVF International Conference on Computer Vision}, pages 3295--3306, 2023.

\bibitem{wang2024vidu4d}
Yikai Wang, Xinzhou Wang, Zilong Chen, Zhengyi Wang, Fuchun Sun, and Jun Zhu.
\newblock Vidu4d: Single generated video to high-fidelity 4d reconstruction with dynamic gaussian surfels.
\newblock {\em arXiv preprint arXiv:2405.16822}, 2024.

\bibitem{4dgs}
Guanjun Wu, Taoran Yi, Jiemin Fang, Lingxi Xie, Xiaopeng Zhang, Wei Wei, Wenyu Liu, Qi Tian, and Xinggang Wang.
\newblock 4d gaussian splatting for real-time dynamic scene rendering.
\newblock In {\em Proceedings of the IEEE/CVF Conference on Computer Vision and Pattern Recognition}, pages 20310--20320, 2024.

\bibitem{nhr}
Minye Wu, Yuehao Wang, Qiang Hu, and Jingyi Yu.
\newblock Multi-view neural human rendering.
\newblock In {\em Proceedings of the IEEE/CVF Conference on Computer Vision and Pattern Recognition}, pages 1682--1691, 2020.

\bibitem{xiao2024bridging}
Yuting Xiao, Xuan Wang, Jiafei Li, Hongrui Cai, Yanbo Fan, Nan Xue, Minghui Yang, Yujun Shen, and Shenghua Gao.
\newblock Bridging 3d gaussian and mesh for freeview video rendering.
\newblock {\em arXiv preprint arXiv:2403.11453}, 2024.

\bibitem{xu20244k4d}
Zhen Xu, Sida Peng, Haotong Lin, Guangzhao He, Jiaming Sun, Yujun Shen, Hujun Bao, and Xiaowei Zhou.
\newblock 4k4d: Real-time 4d view synthesis at 4k resolution.
\newblock In {\em Proceedings of the IEEE/CVF Conference on Computer Vision and Pattern Recognition}, pages 20029--20040, 2024.

\bibitem{yang2024deformable}
Ziyi Yang, Xinyu Gao, Wen Zhou, Shaohui Jiao, Yuqing Zhang, and Xiaogang Jin.
\newblock Deformable 3d gaussians for high-fidelity monocular dynamic scene reconstruction.
\newblock In {\em Proceedings of the IEEE/CVF Conference on Computer Vision and Pattern Recognition}, pages 20331--20341, 2024.

\bibitem{yang2023real}
Zeyu Yang, Hongye Yang, Zijie Pan, Xiatian Zhu, and Li Zhang.
\newblock Real-time photorealistic dynamic scene representation and rendering with 4d gaussian splatting.
\newblock {\em arXiv preprint arXiv:2310.10642}, 2023.

\bibitem{yao2018mvsnet}
Yao Yao, Zixin Luo, Shiwei Li, Tian Fang, and Long Quan.
\newblock Mvsnet: Depth inference for unstructured multi-view stereo.
\newblock In {\em Proceedings of the European conference on computer vision (ECCV)}, pages 767--783, 2018.

\bibitem{yu2024gsdf}
Mulin Yu, Tao Lu, Linning Xu, Lihan Jiang, Yuanbo Xiangli, and Bo Dai.
\newblock Gsdf: 3dgs meets sdf for improved rendering and reconstruction.
\newblock {\em arXiv preprint arXiv:2403.16964}, 2024.

\bibitem{Yu2024GOF}
Zehao Yu, Torsten Sattler, and Andreas Geiger.
\newblock Gaussian opacity fields: Efficient high-quality compact surface reconstruction in unbounded scenes.
\newblock {\em arXiv:2404.10772}, 2024.

\bibitem{lpips}
Richard Zhang, Phillip Isola, Alexei~A Efros, Eli Shechtman, and Oliver Wang.
\newblock The unreasonable effectiveness of deep features as a perceptual metric.
\newblock In {\em CVPR}, 2018.

\end{thebibliography}
}

\end{document}